\documentclass[10pt,twocolumn,letterpaper]{article}

\newif\ifarxiv
\arxivtrue   

\ifarxiv
  \usepackage[pagenumbers]{cvpr}
  \newcommand{\teaserwidth}{0.875\textwidth}
\else
  \usepackage[review]{cvpr}
  \newcommand{\teaserwidth}{\textwidth}
\fi

\usepackage{cuted} 

\usepackage{currfile} 

\usepackage{caption} 

\usepackage{titletoc}

\usepackage{colortbl,multirow,booktabs}

\definecolor{tableone}{rgb}{0.553, 0.749, 0.886}
\definecolor{tabletwo}{rgb}{0.733, 0.851, 0.937}
\definecolor{tablethree}{rgb}{0.851, 0.918, 0.969}

\newcommand{\best}{\cellcolor{tableone}}
\newcommand{\sbest}{\cellcolor{tabletwo}}
\newcommand{\tbest}{\cellcolor{tablethree}}

\renewcommand{\paragraph}[1]{\vspace{.5em}\noindent\textbf{#1.}}

\usepackage{xspace}
\newcommand{\paper}{FlowSplat\xspace}

\definecolor{cvprblue}{rgb}{0.21,0.49,0.74}
\usepackage[pagebackref,breaklinks,colorlinks,allcolors=cvprblue]{hyperref}

\def\confName{CVPR}
\def\confYear{2027}

\title{FlowSplat: Global Gaussian Allocation for Sparse-View Reconstruction\\ Using Normalizing Flows}

\author{
    Guy Levy$^{1}$\qquad
    Dana Cohen-Bar$^{1}$\qquad
    Dor Verbin$^{2}$\qquad
    Or Patashnik$^{1}$\qquad
    Daniel Cohen-Or$^{1}$\\[0.75em]
    $^{1}$Tel Aviv University\qquad
    $^{2}$Google DeepMind
}

\begin{document}
\maketitle
\begin{strip}
    \centering
    \includegraphics[width=\teaserwidth]{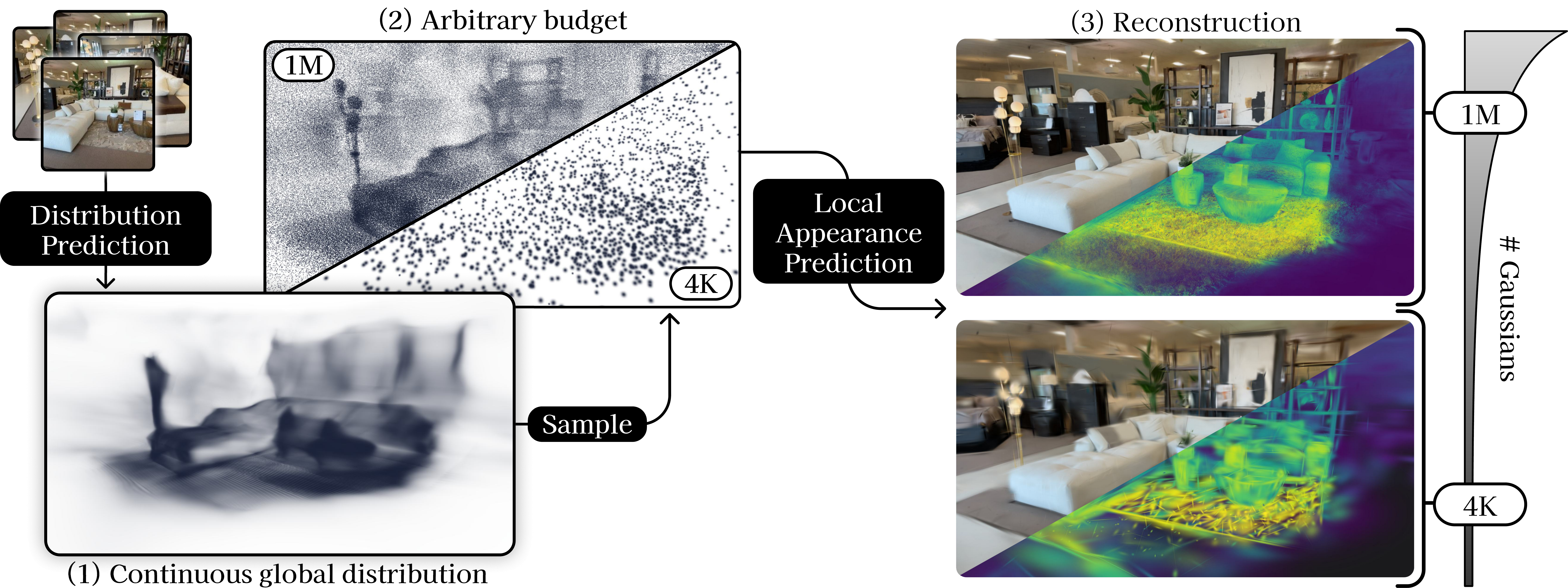}
    \captionsetup{hypcap=false}
    \captionof{figure}{
        \textbf{FlowSplat} reconstructs 3DGS scenes from sparse views by modeling primitive positions as samples from a continuous, scene-conditioned spatial distribution \textbf{(1)}.
        This representation supports sampling an arbitrary number of primitives \textbf{(2)}.
        A point Transformer then predicts the remaining Gaussian attributes from local image features and neighboring primitives,
        producing reconstructions at different levels of detail, shown here with 4K and 1M Gaussians, alongside color-coded visualizations of their spatial density \textbf{(3)}.
    }
    \captionsetup{hypcap=true}
    \label{fig:teaser}
\end{strip}

\begin{abstract}
Gaussian Splatting has become a widely adopted 3D scene representation, combining high visual fidelity with fast rendering.
Although the local appearance of primitives is relatively straightforward to optimize, allocating a finite set of primitives across an entire scene is a challenging global optimization problem. Per-scene optimization techniques typically address this through heuristics for initialization, densification, and pruning.
In feed-forward reconstruction, where these heuristics are difficult to apply, primitive allocation often relies on pixel-aligned prediction or decoders with learned queries that have fixed budgets.
We introduce \textbf{\paper}, a feed-forward scene reconstruction model that learns to predict a global primitive allocation by modeling a scene-conditioned continuous distribution over 3D space. We represent this distribution using normalizing flows, trained end-to-end from rendering losses without direct 3D supervision.
By sampling primitive positions from a continuous distribution, \paper decouples the primitive count from the model architecture, allowing a single model to operate across a wide range of inference-time sampling budgets.
\paper achieves state-of-the-art reconstruction quality with fewer primitives than prior methods, and further improves with larger primitive counts at inference time, without additional training.
\end{abstract}
\section{Introduction}
\label{sec:intro}

3D Gaussian Splatting (3DGS)~\cite{3dgs} has rapidly become a standard representation for novel view synthesis, combining high visual fidelity with real-time rasterization.
Because differentiable rasterization provides gradients only for existing primitives, it enables refining their local appearance and geometry but offers no inherent mechanism for global primitive allocation.
In conventional per-scene optimization~\cite{3dgs,mcmc3dgs}, this global allocation is not modeled explicitly. Rather, primitives are distributed progressively through heuristics to prune redundant primitives and densify them where more primitives are needed.
In feed-forward reconstruction, this challenge becomes particularly pronounced, as the model must infer where and how densely to allocate primitives from sparse input views, without the iterative refinement available in per-scene optimization.

Existing feed-forward methods typically resolve this allocation problem implicitly through their architectural design.
Pixel-aligned approaches~\cite{pixelsplat,mvsplat,depthsplat,idesplat} place primitives along input camera rays, tying both 
their spatial distribution and count to the image grids, creating a mismatch between representational capacity and scene
complexity.
Query-based decoders~\cite{tokengs,querysplat,locusgs} allow more flexible primitive allocation, but are limited by fixed primitive budgets and often underutilize their predicted primitives.
Alternatively, generative models~\cite{diffgs,freerange,nova3r,surflo} provide global allocation but depend on curated 3D targets and computationally expensive training data~\cite{surflo}.

To overcome these limitations, we introduce \textbf{\paper}, a feed-forward reconstruction model that learns a scene-conditioned
continuous probability distribution for global primitive allocation (\cref{fig:teaser}).
Our formulation builds on the recent insight that primitive allocation can be globally optimized directly from rendering 
losses, as demonstrated in the per-scene setting~\cite{egs}.
We realize this formulation using normalizing flows~\cite{normalizingflows,realnvp,papamakarios2021nf}, whose 
invertibility enables efficient sampling of primitive positions and exact likelihood evaluation for rendering-guided 
optimization without direct 3D supervision.
After sampling primitive positions from the learned distribution, a point Transformer~\cite{pointtransformer,ptv3} predicts the 
remaining Gaussian attributes using local multi-view image features and neighboring primitive locations.
Building on recent Transformer-based autoregressive architectures~\cite{tarflow,starflow}, we adapt normalizing flows to 3D space 
using expressive neural spline transformations~\cite{splines}, while improving computational efficiency through 
coarse-to-fine spatial processing.

\paper achieves state-of-the-art reconstruction quality using significantly fewer primitives than prior methods.
Moreover, by decoupling primitive count from model parameters and input resolution, \paper scales to substantially
larger budgets at inference, with reconstruction quality continuing to improve well beyond the training range.
\section{Related work}
\label{sec:2_related_work}

\paragraph{Feed-forward 3D Gaussian splatting}
Feed-forward 3DGS reconstruction aims to recover a renderable scene from images without optimizing each scene separately.
Early approaches adopted pixel-aligned prediction, placing Gaussians along input camera rays using estimated depth~\cite{pixelsplat,mvsplat,depthsplat,gslrm}.
Subsequent methods extended this formulation with stronger multi-view feature aggregation and local 3D interactions between predictions~\cite{longlrm,ilrm,resplat}.

\paragraph{Decoupling primitive count from pixels}
To address grid-aligned redundancy, several methods merge overlapping predictions~\cite{freesplat,ggn,anysplat} or selectively expand local capacity~\cite{offthegrid,splatweaver,gendensification,atsplat}.
Others adjust the primitive budget at inference through scoring, selection, or compression~\cite{ecosplat,f4splat,zipsplat,pocketsplat,sparsesplat}.
Recent query-based decoders instead predict positions using scene-level queries~\cite{tokengs,querysplat,globalsplat,c3g,locusgs}, but typically use fixed primitive budgets and show lower primitive utilization.

\paragraph{Primitive allocation as learned distributions}
Per-scene methods have improved on the densification and pruning rules of 3DGS~\cite{3dgs,revisingadc,taming3dgs} and introduced probabilistic allocation~\cite{mcmc3dgs}.
Recently, Eulerian Gaussian Splatting (EGS)~\cite{egs} presented an effective heuristic-free per-scene approach by modeling primitive positions as samples from a normalized probability density function using an 
image-level score-function estimator that enables rendering gradients to optimize spatial allocation.
Density-Sampled Gaussians (DeG)~\cite{deg} extends probabilistic allocation to feed-forward object generation using a hierarchical octree representation trained with both 3D supervision and rendering losses.
In contrast, \paper learns a resolution-free, scene-conditioned spatial distribution directly from sparse posed images, using rendering supervision alone.
We adopt the EGS estimator for rendering-guided learning, which requires a density model supporting both sampling and tractable likelihood evaluation.

\paragraph{Normalizing flows for images and 3D}
Normalizing flows have long been used for density estimation and generative modeling~\cite{normalizingflows,realnvp,papamakarios2021nf}.
They support sampling from continuous distributions, while their invertible transformations and tractable likelihoods enable likelihood-based training.
These models have been applied across a wide range of domains, including 3D point modeling~\cite{pointflow,dpfnet,softflow},
image-conditioned scene representations~\cite{lasernv}, and importance sampling for rendering~\cite{nis}.
More recently, TARFlow and STARFlow~\cite{tarflow,starflow} demonstrated that normalizing flows can be capable generative models by leveraging scalable Transformer-based autoregressive architectures for image generation.
\paper adapts these architectures to model the spatial distribution of Gaussian primitives,
leveraging tractable log-density evaluation to learn their allocation directly from rendering losses.
\section{Method}
\label{sec:3_method}
\begin{figure*}[t]
    \centering
    \includegraphics[width=\textwidth]{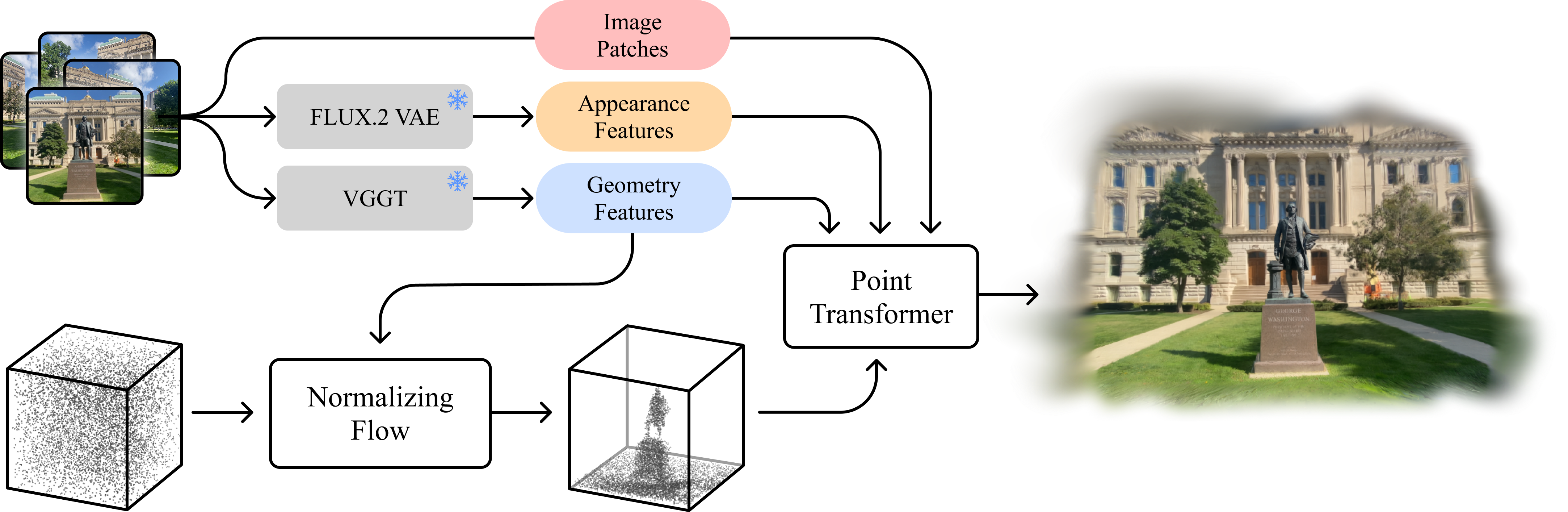} 
    \caption{
        \textbf{Overview of \paper.}
        Given sparse posed input images, frozen VGGT and FLUX.2 VAE models extract geometry and appearance features, respectively.
        A normalizing flow, conditioned on the geometry features, transforms samples from a base distribution into Gaussian positions.
        A point Transformer then predicts the remaining Gaussian attributes using the sampled positions, their spatial configuration, geometry and appearance features, and raw image patches, producing a complete 3D Gaussian representation.
    }
    \label{fig:method_overview}
\end{figure*}

Given a sparse set of posed images, our framework reconstructs a
3D scene by sampling primitive positions from a learned spatial
distribution and predicting their local visual attributes, as
illustrated in \cref{fig:method_overview}.
We first introduce our probabilistic scene representation
(\cref{sec:method_representation}),
then the scene-conditioned normalizing flow that
models this distribution (\cref{sec:method_flow}),
followed by the local attribute network (\cref{sec:method_attributes}),
and finally the training objective (\cref{sec:method_training_objective}).

\subsection{Probabilistic scene representation}
\label{sec:method_representation}

Our input consists of $V$ posed context images
$\mathcal{C}=\{(I_v,\pi_v)\}_{v=1}^{V}$.
We represent the spatial allocation of Gaussian primitives by a
scene-conditioned probability density function
$p_\theta(\boldsymbol{\mu}|\mathcal{C})$.
For a primitive budget $N$, Gaussian means are sampled independently as:
\begin{equation}
    \boldsymbol{\mu}_i
    \overset{\mathrm{i.i.d.}}{\sim}
    p_\theta(\boldsymbol{\mu}|\mathcal{C}),
    \qquad i=1,\ldots,N .
    \label{eq:primitive_sampling}
\end{equation}
Denoting the sampled set of positions by
$\mathcal{M}=\{\boldsymbol{\mu}_i\}_{i=1}^{N}$,
a deterministic attribute network predicts the remaining Gaussian parameters,
namely opacity, scale, rotation, and (view-dependent) color,
conditioned on the context views and the spatial neighborhood of each
primitive:
\begin{equation}
    \mathbf{a}_i =
    A_\phi(\boldsymbol{\mu}_i | \mathcal{M},\mathcal{C}),
    \qquad i=1,\ldots,N .
    \label{eq:attribute_prediction}
\end{equation}
The resulting 3D scene can be written as
$\mathcal{G}_N=\{(\boldsymbol{\mu}_i,\mathbf{a}_i)\}_{i=1}^{N}$.
Since primitive positions are sampled independently from
$p_\theta(\boldsymbol{\mu}|\mathcal{C})$, their number $N$ is a sampling
budget that can be varied at inference time without changing the model.

Given the sampled positions, their local visual attributes can be learned
directly through the differentiable rasterizer.
Learning the spatial distribution, however, requires propagating the rendering
loss back to the probability assigned to each sampled position.
This motivates a density model that supports both efficient sampling and
tractable likelihood evaluation at arbitrary spatial locations.

\subsection{Global primitive allocation}
\label{sec:method_flow}

We parameterize $p_\theta(\boldsymbol{\mu}|\mathcal{C})$ with a
scene-conditioned autoregressive normalizing flow~\cite{maf}.
Given a sample $\mathbf{z}$ from a base distribution $p_0$, the flow maps
$\mathbf{z}$ to a primitive position $\boldsymbol{\mu}$ through an invertible
transformation:
\begin{equation}
    \boldsymbol{\mu}
    = T_\theta(\mathbf{z}|\mathcal{C}),
    \qquad
    \mathbf{z}\sim p_0 .
    \label{eq:flow_forward}
\end{equation}

The corresponding density can be evaluated exactly through the
change-of-variables formula:
\begin{equation}
\log p_\theta(\boldsymbol{\mu}|\mathcal{C})
= \log p_0(\mathbf{z}) -
\log\left|
\det\frac{\partial T_\theta(\mathbf{z}|\mathcal{C})}
{\partial\mathbf{z}}
\right|,
\label{eq:flow_likelihood}
\end{equation}
where $\mathbf{z}=T_\theta^{-1}(\boldsymbol{\mu}|\mathcal{C})$.

We define the domain of our spatial distribution as the unit cube $[0,1]^3$ and use a uniform
distribution over this domain as the flow's base distribution $p_0$.
To handle unbounded scenes, we follow EGS~\cite{egs} and use the same spatial contraction function,
adapted from Mip-NeRF 360~\cite{mipnerf360}, which maps unbounded 3D space to the unit cube.
More details on the spatial contraction are provided in the appendix.

\paragraph{Normalizing flow architecture}
We base our flow architecture on TARFlow~\cite{tarflow}, constructing
$T_\theta$ as a composition of $K$ invertible blocks:
\begin{equation}
    T_\theta = T^{K-1}\circ\cdots\circ T^0 .
    \label{eq:flow_composition}
\end{equation}
Within each block, the three spatial coordinates enter as separate tokens and are
transformed autoregressively, where the transformation of each coordinate depends
on the previously transformed coordinates.
This allows the flow to capture spatial dependencies between axes
while keeping the transformations easy to invert and their likelihoods efficient to evaluate.
Block $k$ maps its input
$\mathbf{z}=(z_0,z_1,z_2)$ to $\mathbf{z}'=(z'_0,z'_1,z'_2)$ as
\begin{equation}
\begin{aligned}
z'_0 &= f(z_0;\boldsymbol{\eta}_{k,0}), \\
z'_1 &= f(z_1;\boldsymbol{\eta}_{k,1}(z'_0)), \\
z'_2 &= f(z_2;\boldsymbol{\eta}_{k,2}(z'_0,z'_1)),
\end{aligned}
\label{eq:autoregressive_block}
\end{equation}
where $f$ is a monotonic rational-quadratic spline~\cite{splines},
and $\boldsymbol{\eta}_{k,j}$ are its parameters.
These splines provide expressive, invertible one-dimensional
transformations over our bounded domain (see Appendix).

\begin{figure}[t]
    \centering
    \includegraphics[width=\linewidth]{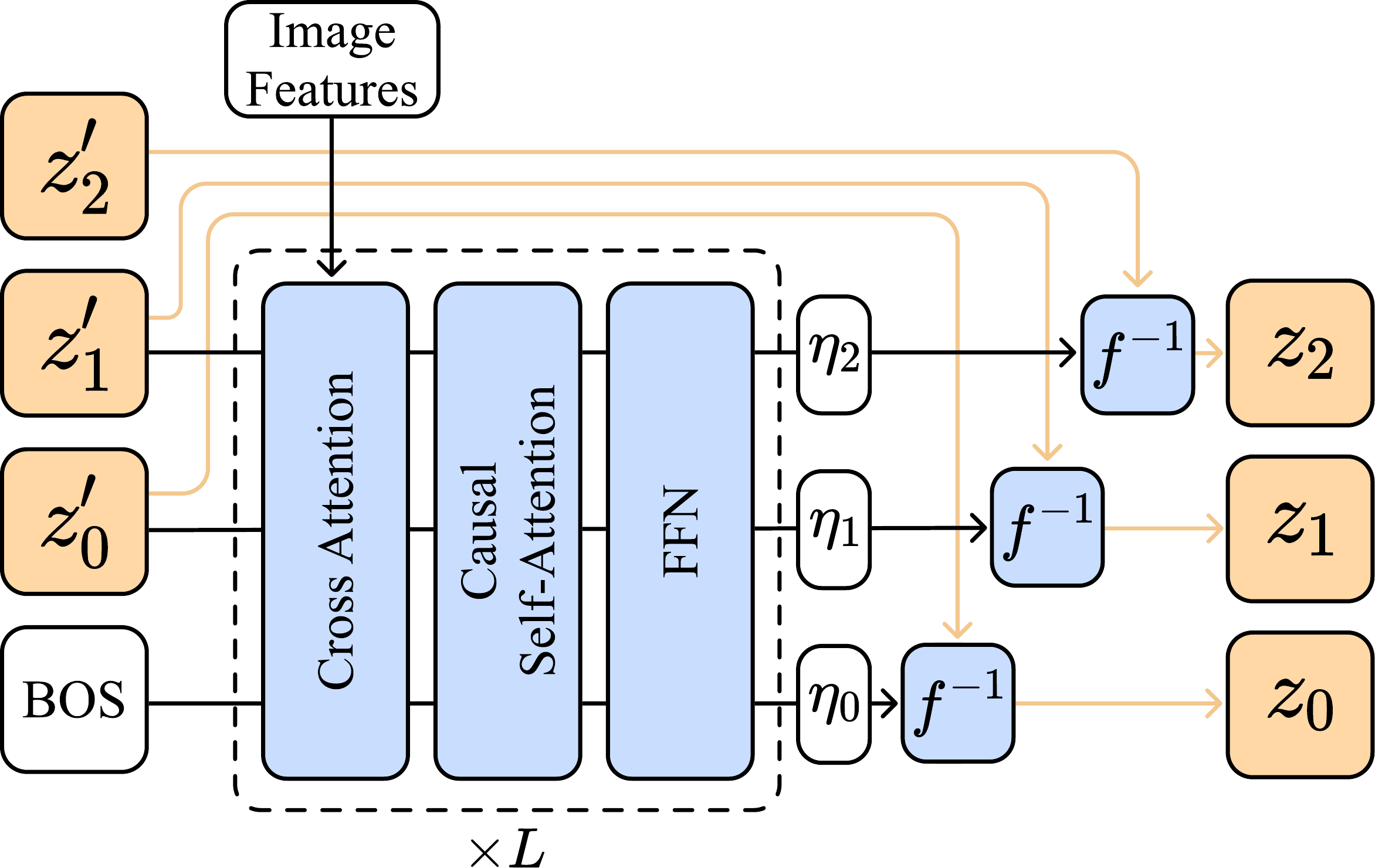}
    \caption{
        \textbf{Autoregressive spline block.}
        In the density-evaluation direction (\mbox{$z'\to z$}), the transformed coordinates
        $(z'_0,z'_1)$ and a BOS token (\textbf{Left}) are processed in parallel through $L$
        Transformer layers (\textbf{Center}) to predict the spline parameters $(\eta_0,\eta_1,\eta_2)$.
        These parameters are used to invert the coordinates in parallel via $f^{-1}$ (\textbf{Right}).
    }
    \label{fig:ar_spline_block}
\end{figure}

Each block predicts the spline parameters ($\eta_{k,j}$) using $L$ Transformer layers,
each consisting of cross-attention to input image features,
causal self-attention across coordinate tokens, and a feed-forward network.
Following the masked autoregressive flow (MAF) formulation~\cite{maf},
the Transformer operates in the inverse (density-evaluation) direction ($z'\to z$).
In this direction, since all transformed coordinates ($z'_i$) are known,
the spline parameters are predicted in parallel, using causal self-attention to
ensure that each transformation depends only on preceding coordinates (\cref{fig:ar_spline_block}).
Sampling, on the other hand, proceeds sequentially, as each transformation
depends on the preceding output coordinates.
For the first coordinate, which has no preceding coordinates, we use
a learned beginning-of-sequence (BOS) token~\cite{starflow}.
The autoregressive structure yields a triangular Jacobian, allowing
its log-determinant to be computed as the sum of one-dimensional
log-derivatives.
We cyclically rotate the coordinate ordering across successive blocks
to avoid privileging any spatial axis.

\begin{figure}[t]
    \centering
    \includegraphics[width=\linewidth]{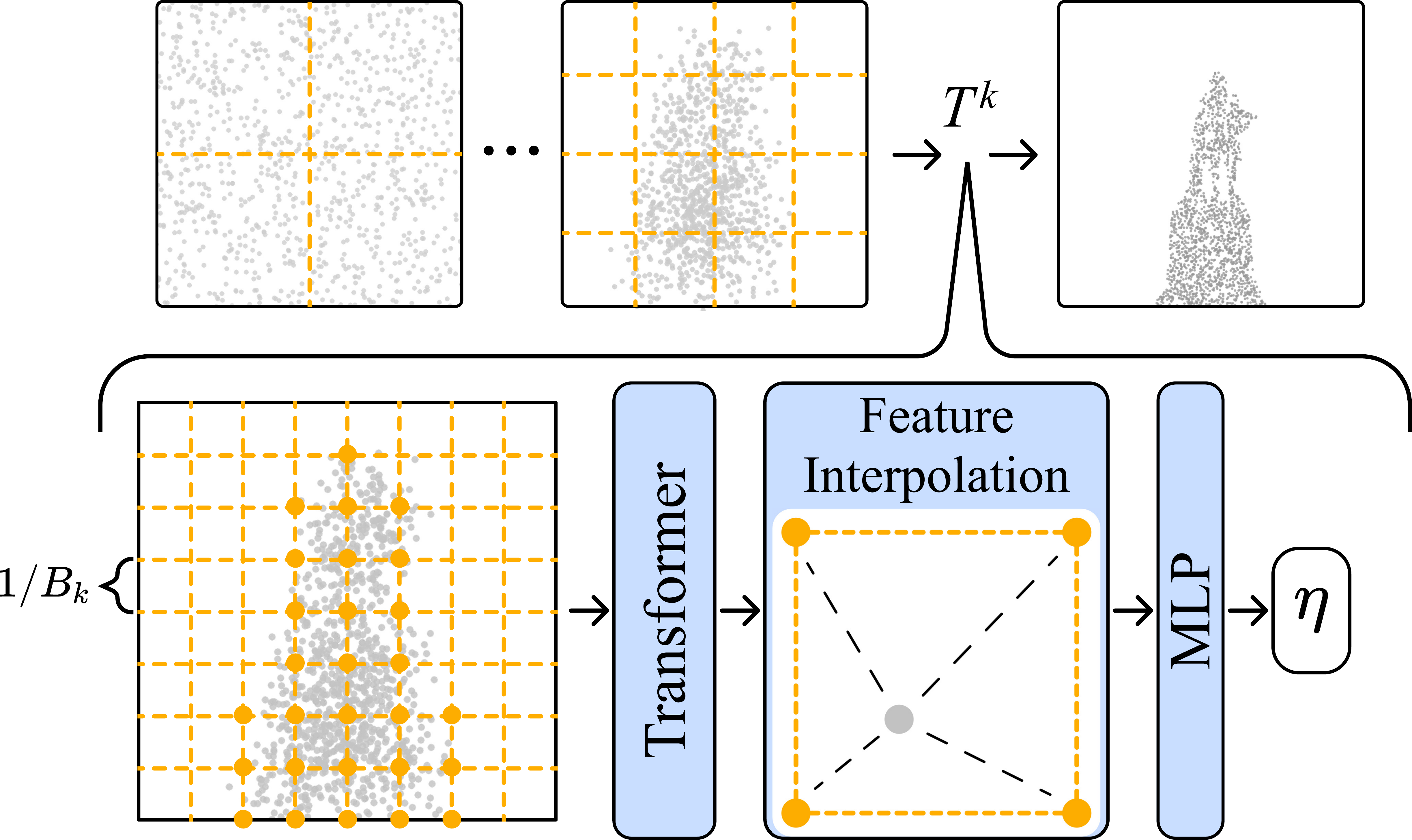}
    \caption{
        \textbf{Progressive prefix quantization.}
        \textbf{Top:} Successive flow blocks transform samples from the base
        distribution into the scene-conditioned density of primitive positions, using
        quantization grids of increasing resolution $B_k$ (orange lines).
        \textbf{Bottom:} At block $T^k$, the Transformer evaluates only a discrete set of
        autoregressive prefix states (orange nodes). For an arbitrary continuous
        position (gray dot), the features of neighboring states are interpolated and
        mapped through an MLP to the spline parameters $\boldsymbol{\eta}$ for the
        continuous position.
        }
    \label{fig:quantization}
\end{figure}

\paragraph{Efficient flow computation}
Evaluating the Transformer independently for every sampled position is prohibitively
expensive at the primitive counts required for scene reconstruction.
Drawing inspiration from hierarchical spatial representations~\cite{instantngp,egs,deg},
we discretize the preceding coordinates at progressively finer resolutions (\cref{fig:quantization}).
At flow block $k$, each preceding coordinate is discretized into $B_k$ values,
forming a one-dimensional grid for the second coordinate and a two-dimensional
grid for the third, while the first uses only the BOS token.
For continuous coordinates, we linearly or bilinearly interpolate the Transformer
features from neighboring grid points to obtain the spline parameters.
We progressively increase $B_k$ across blocks, allowing increasingly fine spatial
modeling while preserving a continuous flow that can be sampled and evaluated at
arbitrary positions.

\paragraph{Scene conditioning}
We condition the flow on multi-scale features extracted from the context
views using a frozen VGGT~\cite{vggt}, fusing features from layers $\{4,11,17,23\}$
through learned softmax weighting as in QuerySplat~\cite{querysplat}.
Following Surflo~\cite{surflo}, we associate each feature token with its
VGGT-predicted 3D position and augment it with a Fourier positional encoding~\cite{fourier}.
Within each flow block, the Transformer cross-attends to these tokens to predict the
spline parameters $\boldsymbol{\eta}_{k,j}$, using the same positional encoding for its spatial queries.
This allows cross-attention to associate spatial queries with scene features
according to their 3D locations.
To match the progressive spatial resolution, at block $k$ we retain only positional
frequencies corresponding to spatial scales no finer than $1/B_k$.

\subsection{Local attribute prediction}
\label{sec:method_attributes}
Given the sampled primitive positions $\mathcal{M}$, we predict the
remaining Gaussian attributes using a point Transformer
that alternates between aggregating multi-view image features and
exchanging information among neighboring primitives.

\paragraph{Multi-view image features}
Our attribute network processes each primitive $\boldsymbol{\mu}_i$ as a separate token,
initialized from a projection of its spatial position and positional encoding.
We project each primitive's position into every context view, where we sample
features from VGGT and the first four stages of a frozen FLUX.2 VAE encoder~\cite{flux2}.
As in the flow model, we fuse VGGT features across layers (using separate learned
mixing weights), while the FLUX features are projected to a common dimension and summed.
We concatenate these features with a local $8 \times 8$ RGB patch from the input
image and the normalized viewing direction.
In each Transformer block, each primitive token cross-attends to these per-view features.

\paragraph{Local primitive interaction}
Since local geometry and appearance in 3DGS depend on the configuration of
neighboring Gaussians, we allow primitives to exchange information during attribute prediction.
Prior feed-forward approaches such as SparseSplat~\cite{sparsesplat} enable
this interaction through explicit $k$-nearest-neighbor graphs, whose
construction becomes increasingly expensive at large primitive counts.
To enable local interaction efficiently, we instead adopt the serialized
attention mechanism of Point Transformer V3~\cite{ptv3}.
We order primitives along locality-preserving Morton and Hilbert
space-filling curves and apply self-attention within local windows
of the resulting sequences. By varying the serialization order across
successive blocks, we allow each primitive to interact with different
spatial neighborhoods without constructing explicit nearest-neighbor graphs.
Following recent serialized point transformers~\cite{litept,ptx}, we encode
relative 3D position within the spatial attention using rotary positional
embeddings~\cite{rope}.

Each attribute block consequently consists of multi-view cross-attention,
serialized local self-attention, and a feed-forward network.
After the final block, the primitive tokens are decoded into opacity, scale,
rotation, and degree-1 spherical-harmonic coefficients.

\begin{table*}[t]
\centering
\footnotesize
\setlength{\tabcolsep}{3.5pt}
\caption{
    \textbf{Quantitative comparison} of novel-view synthesis on the DL3DV-Evaluation
    interpolation cases of QuerySplat~\cite{querysplat}, averaged over small, medium,
    and large splits at $256\times448$ resolution. For our method, we report results
    across three inference-time sampling budgets (14K, 66K, and 262K Gaussians).
{\setlength{\fboxsep}{1pt}
\colorbox{tableone}{\textbf{Best}},
\colorbox{tabletwo}{\underline{second-best}}, and
\colorbox{tablethree}{third-best}}
results are highlighted accordingly.}
\label{tab:qs_interp}

\setlength{\tabcolsep}{3pt}
\begin{tabular}{l cccc cccc cccc}
\toprule
& \multicolumn{4}{c}{2 views}
& \multicolumn{4}{c}{4 views}
& \multicolumn{4}{c}{12 views} \\
\cmidrule(lr){2-5}
\cmidrule(lr){6-9}
\cmidrule(lr){10-13}
Method
& \#G & PSNR$\uparrow$ & SSIM$\uparrow$ & LPIPS$\downarrow$
& \#G & PSNR$\uparrow$ & SSIM$\uparrow$ & LPIPS$\downarrow$
& \#G & PSNR$\uparrow$ & SSIM$\uparrow$ & LPIPS$\downarrow$ \\
\midrule

\multicolumn{13}{l}{\textit{Pixel-aligned}} \\

DepthSplat~\cite{depthsplat}
& 229K & 19.98 & 0.665 & \tbest 0.272
& 459K & 22.52 & 0.763 & \tbest 0.194
& 1.38M & 21.57 & 0.757 & \tbest 0.208 \\

IDESplat~\cite{idesplat}
& 229K & 20.25 & 0.671 & 0.276
& 459K & 23.02 & 0.773 & 0.195
& 1.38M & 21.19 & 0.744 & 0.234 \\

ReSplat (init)~\cite{resplat}
& 14K & 19.53 & 0.640 & 0.343
& 29K & 23.02 & 0.766 & 0.228
& 86K & \tbest 24.21 & \sbest\underline{0.809} & \sbest\underline{0.191} \\

\midrule
\multicolumn{13}{l}{\textit{Query-based}} \\

TokenGS~\cite{tokengs}
& 262K & 19.78 & 0.643 & 0.405
& 262K & 22.39 & 0.724 & 0.334
& 262K & 21.05 & 0.661 & 0.399 \\

LocusGS~\cite{locusgs}
& 262K & 20.36 & 0.671 & 0.368
& 262K & \tbest 23.65 & \tbest 0.774 & 0.278
& 262K & 22.48 & 0.726 & 0.333 \\

\midrule

\multirow{3}{*}{Ours}
& 14K & \tbest 21.01 & \tbest 0.678 & 0.326
& 14K & 23.26 & 0.746 & 0.274
& 14K & 22.50 & 0.711 & 0.315 \\

& 66K & \sbest\underline{22.09} & \sbest\underline{0.727} & \sbest\underline{0.256}
& 66K & \sbest\underline{25.38} & \sbest\underline{0.814} & \sbest\underline{0.187}
& 66K & \sbest\underline{24.95} & \tbest 0.801 & 0.212 \\

& 262K & \best\textbf{22.41} & \best\textbf{0.743} & \best\textbf{0.236}
& 262K & \best\textbf{26.31} & \best\textbf{0.840} & \best\textbf{0.157}
& 262K & \best\textbf{26.36} & \best\textbf{0.846} & \best\textbf{0.162} \\

\bottomrule
\end{tabular}
\end{table*}
\begin{figure*}[t]
    \centering
    \includegraphics[width=\textwidth]{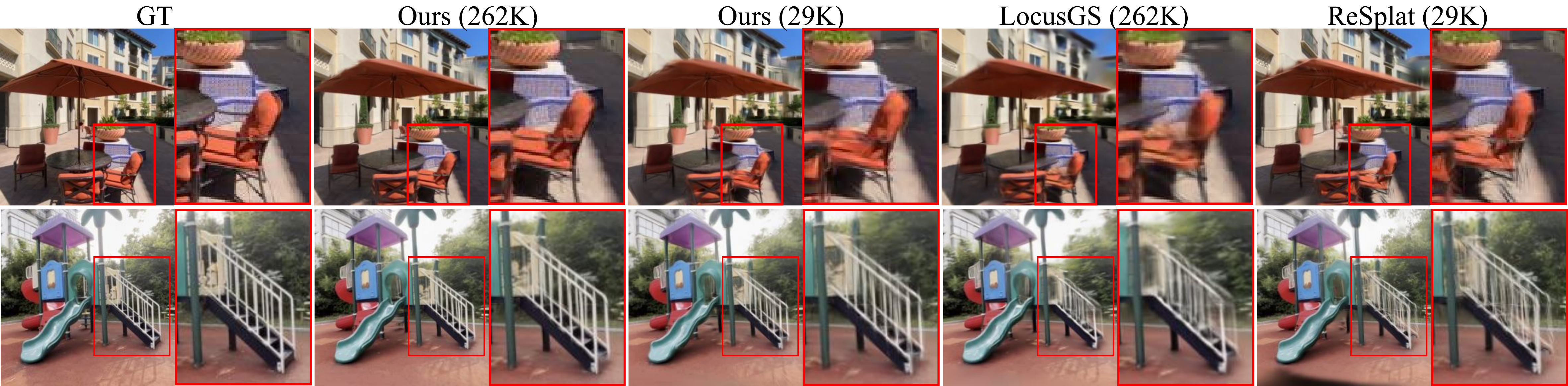} 
    \caption{
        \textbf{Qualitative comparison} on DL3DV-Evaluation using 4 views. The number of Gaussians is indicated next to each method.
    }
    \label{fig:quali}
\end{figure*}

\subsection{Training objective}
\label{sec:method_training_objective}

We train our model end-to-end using a combination of pixelwise reconstruction and perceptual losses:
\begin{equation}
    \mathcal{L} = \mathcal{L}_1 +
    \lambda_{\mathrm{perc}}\mathcal{L}_{\mathrm{perc}},
    \label{eq:render_loss}
\end{equation}
where $\mathcal{L}_{\mathrm{perc}}$ is a VGG-based perceptual loss~\cite{perceptual},
and $\lambda_{\mathrm{perc}}=0.5$ following LVSM~\cite{lvsm}.
While the attribute network is optimized directly through differentiable rasterization~\cite{3dgs},
optimizing the spatial distribution requires relating the rendering loss to the likelihoods of
sampled primitive positions.

To achieve this, we adopt the probabilistic rendering formulation of EGS~\cite{egs}, which uses a
leave-one-out score-function estimator to relate the rendering gradient to the density parameters
through each primitive's contribution to the image.
Under alpha compositing, this yields an opacity-weighted density gradient estimate:
\begin{equation}
    \nabla_\theta \mathcal{L} \approx
    \sum_{i=1}^{N}
    o_i \frac{\partial \mathcal{L}}{\partial o_i}
    \nabla_\theta \log p_\theta(
        \boldsymbol{\mu}_i \mid \mathcal{C}),
    \label{eq:density_gradient}
\end{equation}
where $o_i$ is the opacity of primitive $i$.
Inspired by the reward-clamping strategy of DeG~\cite{deg}, we retain only negative opacity gradients,
replacing $\partial\mathcal{L}/\partial o_i$ with $\min(0,\partial\mathcal{L}/\partial o_i)$.
In effect, this encourages allocating primitives where increasing their opacity would reduce the
rendering loss, without explicitly discouraging allocation where decreasing opacity would help.
We implement the resulting update through an opacity surrogate using standard automatic
differentiation without modifying the rasterizer (see Appendix for details).
To maintain exploration of low probability regions, we additionally employ defensive sampling, drawing $20\%$ of
primitives uniformly from the spatial domain during training.

\section{Experiments}
\label{sec:4_experiments}

\paragraph{Implementation details}
Our normalizing flow consists of $K=5$ autoregressive spline blocks, with progressive spatial resolutions $B_k\in\{2,4,8,16,32\}$.
Each flow block has $L=2$ Transformer layers with a hidden dimension of 1024 and 16 attention heads.
The attribute network consists of 12 Transformer layers with a hidden dimension of 256 and four attention heads.
For local primitive interaction, we use serialized attention over windows of 256 primitives.

\paragraph{Training details}
We train our model on DL3DV-10K~\cite{dl3dv}, excluding the scenes used by DepthSplat~\cite{depthsplat} for evaluation.
Training is performed at $256\times448$ resolution, following the view-sampling protocol of TokenGS~\cite{tokengs}.
We first align the VGGT-predicted cameras and 3D points to the ground-truth camera poses to establish a consistent coordinate frame.
To account for variations in scene scale across DL3DV-10K, we adopt the normalization procedure of Surflo~\cite{surflo}, centering the aligned points at their coordinate-wise 
median and scaling by the median distance to this center.
During training, we sample the primitive count log-uniformly between 32K and 262K to expose the model to different sampling budgets.
Our final model is trained for 150K iterations with a global batch size of 48 using AdamW~\cite{adamw}
with a cosine learning rate schedule.
We use learning rates of $10^{-4}$ for the normalizing flow and feature projections and $3\times10^{-4}$
for the attribute network, applying weight decay of 0.05 only to the normalizing flow.

\subsection{Comparisons}

We evaluate \paper on DL3DV-Evaluation~\cite{dl3dv}, adopting the comprehensive evaluation protocol of
QuerySplat~\cite{querysplat}, which covers varying numbers of input views and interpolation window sizes
 (\cref{tab:qs_interp}).
Our main comparison includes recent pixel-aligned and query-based feed-forward 3DGS reconstruction methods,
evaluated using their released checkpoints.
ReSplat~\cite{resplat} is evaluated using its initial feed-forward prediction without iterative refinement.
In the appendix, we provide additional comparisons to recent methods without released
checkpoints following the evaluation protocol of DepthSplat~\cite{depthsplat}.
We also report cross-dataset generalization and extrapolation results,
and square-frame comparisons to QuerySplat.
Compared to existing feed-forward methods, \paper achieves state-of-the-art reconstruction quality with
substantially fewer primitives, while benefiting further from larger sampling budgets.
With only 66K Gaussians, our method outperforms all baselines in PSNR across all
view counts, using approximately four times fewer primitives than query-based methods.

\subsection{Ablations}
We examine how the main design choices of our framework contribute to reconstruction quality and generalization across primitive budgets.
All ablated models are trained for 30K iterations with a global batch size of 32 on the first six subsets of DL3DV-10K.
Additional architectural ablations are provided in the appendix.

\paragraph{Learned primitive allocation}
A central aspect of our approach is learning where to allocate primitives rather than relying directly on predicted geometry.
To evaluate its contribution, we replace the normalizing flow with positions sampled from VGGT-predicted geometry and train the attribute network independently.
Compared with the baseline, our learned allocation improves reconstruction quality across all three metrics (\cref{tab:ablations}).

\paragraph{Neural spline transformations}
A key part of adapting TARFlow~\cite{tarflow} to model the spatial distribution of primitive
positions is replacing its affine transformations with more expressive neural splines.
To assess the contribution of this change, we train a variant of our model using TARFlow's original
affine transformations, while keeping the remaining architecture unchanged.
The results demonstrate that the spline-based model consistently outperforms
its affine counterpart across all three reconstruction metrics.

\paragraph{FLUX.2 VAE features}
Removing FLUX.2 VAE~\cite{flux2} features from the attribute network while retaining the VGGT features and other inputs leads to a smaller but consistent reduction in reconstruction quality.

\paragraph{Variable-budget training}
To examine the effect of variable-budget training, we compare against
a model trained with a fixed budget of  110K primitives, matching the
average count used to train our model. While both achieve comparable quality within 
the training budget range, variable-budget training generalizes better
to substantially smaller and larger primitive counts.

\subsection{Further analysis and applications}

\begin{figure}[t]
    \centering
    \includegraphics[width=\linewidth]{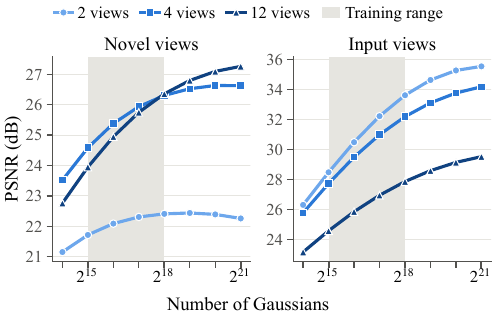}
    \caption{
        \textbf{Inference-time primitive budget scaling.}
        Reconstruction quality scaling as a function of inference-time budget on
        novel (interpolation) views (\textbf{Left}), and on the input views (\textbf{Right}).
    }
    \label{fig:budget_scaling}
\end{figure}

\paragraph{Inference-time primitive budget scaling}
Despite being trained with budgets between 32K and 262K primitives, our model continues to improve with up to 2M primitives, nearly an
order of magnitude beyond the top end of our training range (\cref{fig:budget_scaling}).
Interestingly, this scaling behavior depends strongly on the number of input views.
With only two views, novel-view quality saturates and eventually degrades as the budget increases, whereas with twelve views,
reconstruction quality continues to improve even at the largest budgets.
In contrast, reconstruction quality on the input views continues to improve across all view counts, suggesting that additional primitives
can increasingly fit the observed images without necessarily improving novel-view reconstruction when observations are sparse.

\begin{table}[t]
\centering
\footnotesize
\setlength{\tabcolsep}{4pt}
\caption{\textbf{Ablations} on DL3DV-Evaluation using the 4-view interpolation cases of QuerySplat. \#G denotes the number of Gaussian primitives per scene. The best result for each primitive count is shown in \textbf{bold}.}
\label{tab:ablations}
\begin{tabular}{llccc}
\toprule
Variant & \#G & PSNR$\uparrow$ & SSIM$\uparrow$ & LPIPS$\downarrow$ \\
\midrule
Ours (full model) & 16K & \textbf{22.64} & \textbf{0.726} & \textbf{0.290} \\
 & 110K & 24.87 & 0.805 & 0.195 \\
 & 1M & \textbf{25.76} & \textbf{0.834} & \textbf{0.165} \\
\midrule
Placement at VGGT points & 110K & 23.27 & 0.762 & 0.231 \\
\midrule
Ours w/ affine transforms & 110K & 24.35 & 0.790 & 0.210 \\
\midrule
Ours w/o FLUX.2 VAE features & 110K & 24.58 & 0.797 & 0.208 \\
\midrule
Fixed count (110K) training & 16K & 22.00 & 0.710 & 0.307 \\
 & 110K & \textbf{24.94} & \textbf{0.807} & \textbf{0.193} \\
 & 1M & 25.64 & 0.832 & 0.166 \\
\bottomrule
\end{tabular}
\end{table}

\begin{table}[t]
\centering
\footnotesize
\setlength{\tabcolsep}{2.5pt}
\caption{
    \textbf{Geometric agreement on DL3DV-Evaluation.}
    We compare predicted Gaussian positions against reference geometry.
    We report accuracy (Acc.), completeness (Comp.), Chamfer distance (CD), precision (P), recall (R), and F1 score.
    Best and second-best results are shown in \textbf{bold} and \underline{underlined}, respectively.
}
\label{tab:geometry}
\resizebox{\columnwidth}{!}{%
\begin{tabular}{l cccccc}
\toprule
Method & Acc.$\downarrow$ & Comp.$\downarrow$ & CD$\downarrow$ & P$\uparrow$ & R$\uparrow$ & F1$\uparrow$ \\
\midrule
\multicolumn{7}{l}{\textit{Pixel-aligned}} \\
DepthSplat~\cite{depthsplat} & \underline{0.669} & 1.309 & 0.989 & 0.817 & 0.624 & 0.700 \\
IDESplat~\cite{idesplat} & 0.681 & 1.342 & 1.011 & \underline{0.818} & 0.627 & 0.702 \\
ReSplat (init)~\cite{resplat} & 0.696 & \underline{1.228} & \textbf{0.962} & 0.814 & \underline{0.664} & \underline{0.724} \\
\midrule
\multicolumn{7}{l}{\textit{Query-based}} \\
TokenGS~\cite{tokengs} & 3.899 & 1.544 & 2.722 & 0.424 & 0.532 & 0.450 \\
LocusGS~\cite{locusgs} & \textbf{0.573} & 1.519 & 1.046 & \textbf{0.859} & 0.599 & 0.699 \\
\midrule
Ours & 0.768 & \textbf{1.164} & \underline{0.966} & 0.796 & \textbf{0.680} & \textbf{0.728} \\
\bottomrule
\end{tabular}}
\end{table}

\begin{figure*}[t]
    \centering
    \includegraphics[width=\textwidth]{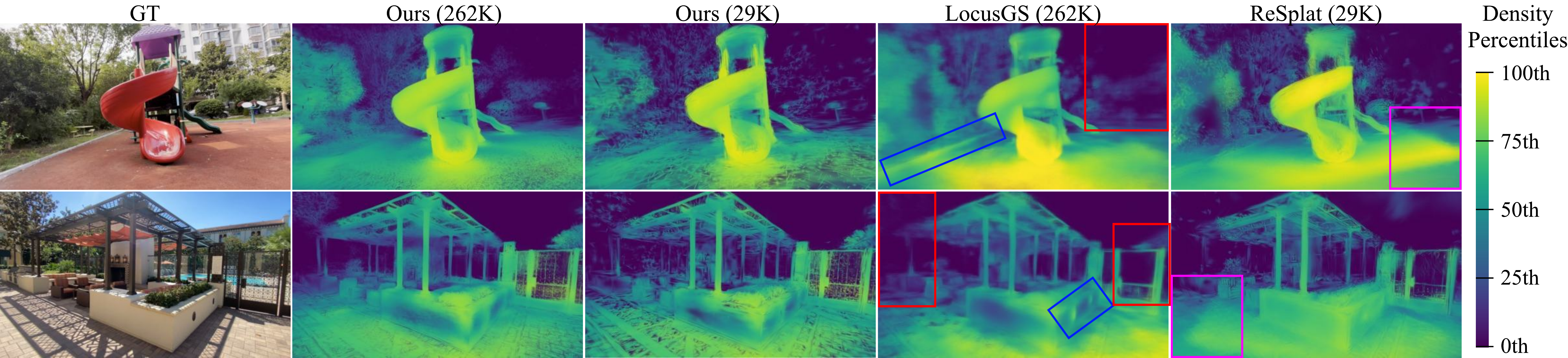} 
    \caption{
        \textbf{Comparative density visualization.}
        We estimate local density by counting neighboring primitives within a radius of $0.1$ times
        the camera spread, coloring each primitive by its percentile within each reconstruction.
        Compared to \paper, LocusGS~\cite{locusgs} shows greater local density variations
        (\textbf{blue}) and reduced coverage of distant structures (\textbf{red}). ReSplat~\cite{resplat} exhibits
        content-independent coverage and view boundary artifacts (\textbf{magenta}).
    }
    \label{fig:density_quali}
\end{figure*}

\paragraph{Learned allocation and primitive utilization}
Beyond reconstruction quality, we analyze the spatial distribution of our sampled primitives through geometric accuracy, qualitative
density visualizations, and primitive utilization.

Following Surflo~\cite{surflo}, we obtain pseudo-ground-truth geometry for DL3DV-Evaluation by
reconstructing each scene from its full set of views using Gaussian Wrapping~\cite{gaussianwrapping}.
We then sample reference points from the extracted meshes and evaluate the geometric
agreement between the predicted Gaussian positions of each method
and the reference samples that lie within the cumulative frustum of the interpolation window (see Appendix).
As shown in~\cref{tab:geometry}, our method achieves the best completeness, recall, and F1 score, alongside the second-best Chamfer distance.
While LocusGS~\cite{locusgs} and the pixel-aligned baselines achieve higher accuracy and precision, our method provides more complete geometric coverage,
suggesting a different trade-off between local precision and scene completeness.

We further visualize the spatial allocation of primitives using color-coded local density 
estimates in~\cref{fig:density_quali}.
Compared to \paper, the query-based LocusGS achieves better local contrast around high-frequency 
elements closer to the input cameras, while under-allocating similarly detailed areas further away.
Being pixel-aligned, ReSplat shows content-independent coverage of the scene, with noticeable artifacts
around the boundaries of input views.

\begin{figure}[t]
    \centering
    \includegraphics[width=\linewidth]{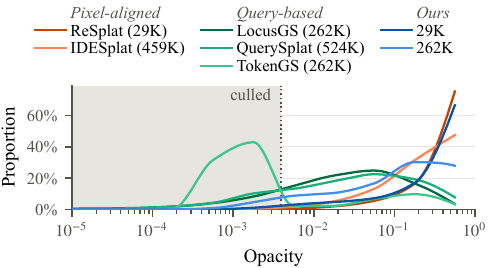}
    \caption{\textbf{Opacity distribution comparison.}
        Our method (\textbf{blue}) provides a substantially lower proportion of culled
        primitives than query-based methods (\textbf{green}),
        comparable to pixel-aligned baselines (\textbf{orange}).
        The shaded region indicates primitives below the opacity culling threshold.
    }
    \label{fig:utilization_dist}
\end{figure}

Lastly, we examine primitive utilization through opacity distributions, adopting the
analysis of QuerySplat~\cite{querysplat} (\cref{fig:utilization_dist}).
In the four-view setting, 76.1\% of TokenGS, 21.0\% of QuerySplat,
and 20.4\% of LocusGS primitives fall below the opacity-culling threshold,
compared with only 7.2\% and 1.8\% for our model at 262K and 29K primitives respectively.
Pixel-aligned methods exhibit even lower culling rates, with
1.0\% for IDESplat and 0.2\% for ReSplat.
Overall, \paper provides substantially better primitive utilization than query-based approaches,
without tying allocation to input image grids.

\paragraph{Progressive local refinement}
\begin{figure}[t]
    \centering
    \includegraphics[width=\linewidth]{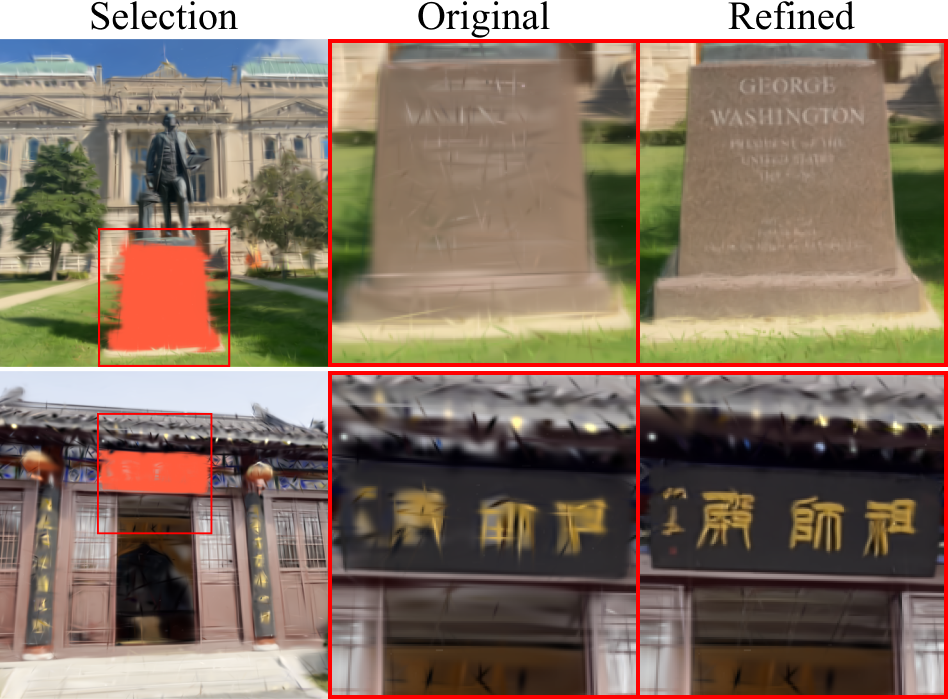}
    \caption{
    \textbf{Progressive local refinement.} By inverting selected primitives into the base distribution and sampling additional
        positions around them, we locally refine existing reconstructions, recovering finer details without reconstructing the entire scene.
    }
    \label{fig:refinement}
\end{figure}

A unique advantage of our normalizing flow formulation is its invertibility,
which enables local refinement of existing reconstructions.
Given a selected region, we map its primitives to the base distribution, sample additional positions
around their latent coordinates, and transform them back to the scene.
The attribute network then predicts the remaining Gaussian parameters, allowing us to increase
primitive density in selected regions without reconstructing the entire scene (\cref{fig:refinement}).
We describe this process in detail in the appendix.
\section{Conclusions and limitations}
\label{sec:5_conclusion}

We introduced a feed-forward reconstruction model that learns global Gaussian allocation through a scene-conditioned continuous distribution.
Our results show the benefit of explicit modeling of global primitive allocation alongside attribute prediction,
and that this allocation can be learned directly from rendering losses.
We demonstrated that modern normalizing flows are an effective framework for this purpose,
enabling likelihood-based training for rendering-guided allocation.

Our approach, while providing better quality compared to previous methods, does not come without cost.
As we discuss in the appendix, \paper is more computationally expensive than most query-based methods.
Future work could explore more efficient sampling as well as sharing more of the computation across primitives during attribute prediction.
Another interesting direction is to extend the learned allocation to dynamic scenes through a continuous distribution over space and time.
\section*{Acknowledgments}
This research was supported in part by the Israel Science Foundation (grants no. 2492/20, 3441/21, and 1473/24), Len Blavatnik and the Blavatnik Family Foundation, and the Bronfman Family Foundation. This research was made possible through a GPU compute resource grant funded by the Association of University Heads, the Council for Higher Education and the AI Research Compute Center.
{
    \small
    \bibliographystyle{ieeenat_fullname}
    \bibliography{main}

\begin{thebibliography}{69}
\providecommand{\natexlab}[1]{#1}
\providecommand{\url}[1]{\texttt{#1}}
\expandafter\ifx\csname urlstyle\endcsname\relax
  \providecommand{\doi}[1]{doi: #1}\else
  \providecommand{\doi}{doi: \begingroup \urlstyle{rm}\Url}\fi

\bibitem[An et~al.(2025)An, Jung, Kim, Kim, Jeon, Han, Fukuda, Narihira, Ko, Kim, Hong, Mitsufuji, and Kim]{c3g}
Honggyu An, Jaewoo Jung, Mungyeom Kim, Chaehyun Kim, Minkyeong Jeon, Jisang Han, Kazumi Fukuda, Takuya Narihira, Hyuna Ko, Junsu Kim, Sunghwan Hong, Yuki Mitsufuji, and Seungryong Kim.
\newblock {C3G}: Learning compact {3D} representations with {2K} {G}aussians.
\newblock \emph{arXiv preprint arXiv:2512.04021}, 2025.

\bibitem[Barron et~al.(2022)Barron, Mildenhall, Verbin, Srinivasan, and Hedman]{mipnerf360}
Jonathan~T. Barron, Ben Mildenhall, Dor Verbin, Pratul~P. Srinivasan, and Peter Hedman.
\newblock {Mip-NeRF} 360: Unbounded anti-aliased neural radiance fields.
\newblock In \emph{CVPR}, pages 5470--5479, 2022.

\bibitem[{Black Forest Labs}(2025)]{flux2}
{Black Forest Labs}.
\newblock {FLUX.2}: Analyzing and enhancing the latent space of {FLUX} -- representation comparison.
\newblock \url{https://bfl.ai/research/representation-comparison}, 2025.

\bibitem[Charatan et~al.(2024)Charatan, Li, Tagliasacchi, and Sitzmann]{pixelsplat}
David Charatan, Sizhe~Lester Li, Andrea Tagliasacchi, and Vincent Sitzmann.
\newblock {pixelSplat}: {3D} {G}aussian splats from image pairs for scalable generalizable {3D} reconstruction.
\newblock In \emph{CVPR}, pages 19457--19467, 2024.

\bibitem[Chen et~al.(2026)Chen, Zheng, Zhang, Vedaldi, and Cremers]{nova3r}
Weirong Chen, Chuanxia Zheng, Ganlin Zhang, Andrea Vedaldi, and Daniel Cremers.
\newblock {NOVA3R}: Non-pixel-aligned visual transformer for amodal {3D} reconstruction.
\newblock In \emph{ICLR}, 2026.

\bibitem[Chen et~al.(2024)Chen, Xu, Zheng, Zhuang, Pollefeys, Geiger, Cham, and Cai]{mvsplat}
Yuedong Chen, Haofei Xu, Chuanxia Zheng, Bohan Zhuang, Marc Pollefeys, Andreas Geiger, Tat-Jen Cham, and Jianfei Cai.
\newblock {MVSplat}: Efficient {3D} {G}aussian splatting from sparse multi-view images.
\newblock In \emph{ECCV}, pages 370--386, 2024.

\bibitem[Cho et~al.(2026)Cho, Cho, Yoo, Lee, and Kim]{atsplat}
In Cho, Jeonghwan Cho, Mijin Yoo, Gim~Hee Lee, and Seon~Joo Kim.
\newblock {ATSplat}: Compact feed-forward {3D} {G}aussian splatting with adaptive token expansion.
\newblock \emph{arXiv preprint arXiv:2607.20417}, 2026.

\bibitem[Dinh et~al.(2017)Dinh, Sohl-Dickstein, and Bengio]{realnvp}
Laurent Dinh, Jascha Sohl-Dickstein, and Samy Bengio.
\newblock Density estimation using {Real NVP}.
\newblock In \emph{ICLR}, 2017.

\bibitem[Durkan et~al.(2019)Durkan, Bekasov, Murray, and Papamakarios]{splines}
Conor Durkan, Artur Bekasov, Iain Murray, and George Papamakarios.
\newblock Neural spline flows.
\newblock In \emph{NeurIPS}, 2019.

\bibitem[Gomez et~al.(2026)Gomez, Gu{\'e}don, Maruani, Gong, and Ovsjanikov]{gaussianwrapping}
Diego Gomez, Antoine Gu{\'e}don, Nissim Maruani, Bingchen Gong, and Maks Ovsjanikov.
\newblock From blobs to spokes: High-fidelity surface reconstruction via oriented {G}aussians.
\newblock In \emph{ECCV}, pages 54--72, 2026.

\bibitem[Gu et~al.(2025)Gu, Chen, Berthelot, Zheng, Wang, Zhang, Dinh, Bautista, Susskind, and Zhai]{starflow}
Jiatao Gu, Tianrong Chen, David Berthelot, Huangjie Zheng, Yuyang Wang, Ruixiang Zhang, Laurent Dinh, Miguel~Angel Bautista, Joshua Susskind, and Shuangfei Zhai.
\newblock {STARFlow}: Scaling latent normalizing flows for high-resolution image synthesis.
\newblock In \emph{NeurIPS}, pages 134122--134158, 2025.

\bibitem[Gu{\'e}don et~al.(2026)Gu{\'e}don, Nakamura, Dufour, Lei, Nishino, and Kanazawa]{surflo}
Antoine Gu{\'e}don, Shu Nakamura, Nicolas Dufour, Jiahui Lei, Ko Nishino, and Angjoo Kanazawa.
\newblock {Surflo}: Consistent {3D} surface flow model with global state.
\newblock In \emph{NeurIPS}, 2026.

\bibitem[Guo et~al.(2026)Guo, Chen, Chen, Fang, and Wang]{pocketsplat}
Wenzhi Guo, Xianda Chen, Dongxuan Chen, Guangchi Fang, and Bing Wang.
\newblock {PocketSplat}: Mobile {G}aussian reconstruction via world-space latent allocation.
\newblock \emph{arXiv preprint arXiv:2610.03192}, 2026.

\bibitem[Itkin et~al.(2026)Itkin, Issachar, Keypur, Chen, Chen, and Benaim]{globalsplat}
Roni Itkin, Noam Issachar, Yehonatan Keypur, Xingyu Chen, Anpei Chen, and Sagie Benaim.
\newblock {GlobalSplat}: Efficient feed-forward {3D} {G}aussian splatting via global scene tokens.
\newblock In \emph{ECCV}, pages 171--189, 2026.

\bibitem[Jiang et~al.(2025)Jiang, Mao, Xu, Lu, Ren, Jin, Xu, Yu, Pang, Zhao, Lin, and Dai]{anysplat}
Lihan Jiang, Yucheng Mao, Linning Xu, Tao Lu, Kerui Ren, Yichen Jin, Xudong Xu, Mulin Yu, Jiangmiao Pang, Feng Zhao, Dahua Lin, and Bo Dai.
\newblock {AnySplat}: Feed-forward {3D} {G}aussian splatting from unconstrained views.
\newblock \emph{ACM TOG}, 44\penalty0 (6):\penalty0 1--16, 2025.

\bibitem[Jin et~al.(2025)Jin, Jiang, Tan, Zhang, Bi, Zhang, Luan, Snavely, and Xu]{lvsm}
Haian Jin, Hanwen Jiang, Hao Tan, Kai Zhang, Sai Bi, Tianyuan Zhang, Fujun Luan, Noah Snavely, and Zexiang Xu.
\newblock {LVSM}: A large view synthesis model with minimal {3D} inductive bias.
\newblock In \emph{ICLR}, 2025.

\bibitem[Johnson et~al.(2016)Johnson, Alahi, and Fei-Fei]{perceptual}
Justin Johnson, Alexandre Alahi, and Li Fei-Fei.
\newblock Perceptual losses for real-time style transfer and super-resolution.
\newblock In \emph{ECCV}, pages 694--711, 2016.

\bibitem[Kang et~al.(2026)Kang, Nam, Yang, Sun, Khamis, Mohamed, and Park]{ilrm}
Gyeongjin Kang, Seungtae Nam, Seungkwon Yang, Xiangyu Sun, Sameh Khamis, Abdelrahman Mohamed, and Eunbyung Park.
\newblock {iLRM}: An iterative large {3D} reconstruction model.
\newblock In \emph{CVPR}, pages 37332--37342, 2026.

\bibitem[Kerbl et~al.(2023)Kerbl, Kopanas, Leimk{\"u}hler, and Drettakis]{3dgs}
Bernhard Kerbl, Georgios Kopanas, Thomas Leimk{\"u}hler, and George Drettakis.
\newblock {3D} {G}aussian splatting for real-time radiance field rendering.
\newblock \emph{ACM TOG}, 42\penalty0 (4):\penalty0 1--14, 2023.

\bibitem[Kheradmand et~al.(2024)Kheradmand, Rebain, Sharma, Sun, Tseng, Isack, Kar, Tagliasacchi, and Yi]{mcmc3dgs}
Shakiba Kheradmand, Daniel Rebain, Gopal Sharma, Weiwei Sun, Yang-Che Tseng, Hossam Isack, Abhishek Kar, Andrea Tagliasacchi, and Kwang~Moo Yi.
\newblock {3D} {G}aussian splatting as {M}arkov chain {M}onte {C}arlo.
\newblock In \emph{NeurIPS}, pages 80965--80986, 2024.

\bibitem[Kim et~al.(2020)Kim, Lee, Kang, Lee, and Kim]{softflow}
Hyeongju Kim, Hyeonseung Lee, Woo~Hyun Kang, Joun~Yeop Lee, and Nam~Soo Kim.
\newblock {SoftFlow}: Probabilistic framework for normalizing flow on manifolds.
\newblock In \emph{NeurIPS}, pages 16388--16397, 2020.

\bibitem[Kim et~al.(2026)Kim, Kim, Bae, Joo, and Kim]{f4splat}
Injae Kim, Chaehyeon Kim, Minseong Bae, Minseok Joo, and Hyunwoo~J. Kim.
\newblock {F4Splat}: Feed-forward predictive densification for feed-forward {3D} {G}aussian splatting.
\newblock In \emph{ECCV}, pages 1--19, 2026.

\bibitem[Klokov et~al.(2020)Klokov, Boyer, and Verbeek]{dpfnet}
Roman Klokov, Edmond Boyer, and Jakob Verbeek.
\newblock Discrete point flow networks for efficient point cloud generation.
\newblock In \emph{ECCV}, pages 694--710, 2020.

\bibitem[Li et~al.(2026{\natexlab{a}})Li, Liu, Hu, Qiao, and Dou]{locusgs}
Wenyu Li, Sidun Liu, Tongrui Hu, Peng Qiao, and Yong Dou.
\newblock {LocusGS}: Spatially grounded tokens for feed-forward {3D} {G}aussian splatting.
\newblock \emph{arXiv preprint arXiv:2608.12825}, 2026{\natexlab{a}}.

\bibitem[Li et~al.(2026{\natexlab{b}})Li, Shen, Zhang, Ye, Wu, Hao, Zhang, and Liu]{querysplat}
Yinglong Li, Donghui Shen, Xiaoyu Zhang, Zhichao Ye, Hongyu Wu, Aimin Hao, Guofeng Zhang, and Haomin Liu.
\newblock {QuerySplat}: Decoupling geometry and appearance representations in {3DGS} prediction.
\newblock \emph{arXiv preprint arXiv:2608.01186}, 2026{\natexlab{b}}.

\bibitem[Ling et~al.(2024)Ling, Sheng, Tu, Zhao, Xin, Wan, Yu, Guo, Yu, Lu, Li, Sun, Ashok, Mukherjee, Kang, Kong, Hua, Zhang, Benes, and Bera]{dl3dv}
Lu Ling, Yichen Sheng, Zhi Tu, Wentian Zhao, Cheng Xin, Kun Wan, Lantao Yu, Qianyu Guo, Zixun Yu, Yawen Lu, Xuanmao Li, Xingpeng Sun, Rohan Ashok, Aniruddha Mukherjee, Hao Kang, Xiangrui Kong, Gang Hua, Tianyi Zhang, Bedrich Benes, and Aniket Bera.
\newblock {DL3DV-10K}: A large-scale scene dataset for deep learning-based {3D} vision.
\newblock In \emph{CVPR}, pages 22160--22169, 2024.

\bibitem[Liu et~al.(2021)Liu, Tucker, Jampani, Makadia, Snavely, and Kanazawa]{acid}
Andrew Liu, Richard Tucker, Varun Jampani, Ameesh Makadia, Noah Snavely, and Angjoo Kanazawa.
\newblock Infinite nature: Perpetual view generation of natural scenes from a single image.
\newblock In \emph{ICCV}, pages 14458--14467, 2021.

\bibitem[Long et~al.(2026)Long, Wu, Jiang, Zhang, Ji, and Gu]{idesplat}
Wei Long, Haifeng Wu, Shiyin Jiang, Jinhua Zhang, Xinchun Ji, and Shuhang Gu.
\newblock {IDESplat}: Iterative depth probability estimation for generalizable {3D} {G}aussian splatting.
\newblock In \emph{CVPR}, pages 33248--33258, 2026.

\bibitem[Loshchilov and Hutter(2019)]{adamw}
Ilya Loshchilov and Frank Hutter.
\newblock Decoupled weight decay regularization.
\newblock In \emph{ICLR}, 2019.

\bibitem[Mallick et~al.(2024)Mallick, Goel, Kerbl, Steinberger, Carrasco, and De~La~Torre]{taming3dgs}
Saswat~Subhajyoti Mallick, Rahul Goel, Bernhard Kerbl, Markus Steinberger, Francisco~Vicente Carrasco, and Fernando De~La~Torre.
\newblock Taming {3DGS}: High-quality radiance fields with limited resources.
\newblock In \emph{SIGGRAPH Asia Conference Papers}, pages 1--11, 2024.

\bibitem[McCandlish et~al.(2018)McCandlish, Kaplan, Amodei, and {OpenAI Dota Team}]{gradientsnr}
Sam McCandlish, Jared Kaplan, Dario Amodei, and {OpenAI Dota Team}.
\newblock An empirical model of large-batch training.
\newblock \emph{arXiv preprint arXiv:1812.06162}, 2018.

\bibitem[Moreau et~al.(2026)Moreau, Shaw, Nazarczuk, Shin, Tanay, Zhang, Xu, and P{\'e}rez-Pellitero]{offthegrid}
Arthur Moreau, Richard Shaw, Michal Nazarczuk, Jisu Shin, Thomas Tanay, Zhensong Zhang, Songcen Xu, and Eduardo P{\'e}rez-Pellitero.
\newblock Off the grid: Detection of primitives for feed-forward {3D} {G}aussian splatting.
\newblock In \emph{CVPR}, pages 11756--11766, 2026.

\bibitem[Moreno et~al.(2023)Moreno, Kosiorek, Strathmann, Zoran, Schneider, Winckler, Markeeva, Weber, and Rezende]{lasernv}
Pol Moreno, Adam~R. Kosiorek, Heiko Strathmann, Daniel Zoran, Rosalia~G. Schneider, Bj{\"o}rn Winckler, Larisa Markeeva, Th{\'e}ophane Weber, and Danilo~J. Rezende.
\newblock Laser: Latent set representations for {3D} generative modeling.
\newblock \emph{arXiv preprint arXiv:2301.05747}, 2023.

\bibitem[M{\"u}ller et~al.(2019)M{\"u}ller, McWilliams, Rousselle, Gross, and Nov{\'a}k]{nis}
Thomas M{\"u}ller, Brian McWilliams, Fabrice Rousselle, Markus Gross, and Jan Nov{\'a}k.
\newblock Neural importance sampling.
\newblock \emph{ACM TOG}, 38\penalty0 (5):\penalty0 1--19, 2019.

\bibitem[M{\"u}ller et~al.(2022)M{\"u}ller, Evans, Schied, and Keller]{instantngp}
Thomas M{\"u}ller, Alex Evans, Christoph Schied, and Alexander Keller.
\newblock Instant neural graphics primitives with a multiresolution hash encoding.
\newblock \emph{ACM TOG}, 41\penalty0 (4):\penalty0 1--15, 2022.

\bibitem[Nam et~al.(2025)Nam, Sun, Kang, Lee, Oh, and Park]{gendensification}
Seungtae Nam, Xiangyu Sun, Gyeongjin Kang, Younggeun Lee, Seungjun Oh, and Eunbyung Park.
\newblock Generative densification: Learning to densify {G}aussians for high-fidelity generalizable {3D} reconstruction.
\newblock In \emph{CVPR}, pages 26683--26693, 2025.

\bibitem[Owen(1998)]{scrambling}
Art~B. Owen.
\newblock Scrambling {Sobol'} and {Niederreiter--Xing} points.
\newblock \emph{J. Complexity}, 14\penalty0 (4):\penalty0 466--489, 1998.

\bibitem[Papamakarios et~al.(2017)Papamakarios, Pavlakou, and Murray]{maf}
George Papamakarios, Theo Pavlakou, and Iain Murray.
\newblock Masked autoregressive flow for density estimation.
\newblock In \emph{NeurIPS}, 2017.

\bibitem[Papamakarios et~al.(2021)Papamakarios, Nalisnick, Rezende, Mohamed, and Lakshminarayanan]{papamakarios2021nf}
George Papamakarios, Eric Nalisnick, Danilo~Jimenez Rezende, Shakir Mohamed, and Balaji Lakshminarayanan.
\newblock Normalizing flows for probabilistic modeling and inference.
\newblock \emph{JMLR}, 22\penalty0 (57):\penalty0 1--64, 2021.

\bibitem[Park et~al.(2026)Park, Bui, Gonzalez~Bello, Moon, Oh, and Kim]{ecosplat}
Jongmin Park, Minh-Quan~Viet Bui, Juan~Luis Gonzalez~Bello, Jaeho Moon, Jihyong Oh, and Munchurl Kim.
\newblock {EcoSplat}: Efficiency-controllable feed-forward {3D} {G}aussian splatting from multi-view images.
\newblock In \emph{CVPR}, pages 26010--26020, 2026.

\bibitem[Polansky et~al.(2026)Polansky, Kopanas, Garbin, Zickler, and Verbin]{egs}
Mia~Gaia Polansky, George Kopanas, Stephan Garbin, Todd Zickler, and Dor Verbin.
\newblock {E}ulerian {G}aussian splatting using hashed probability pyramids.
\newblock In \emph{CVPR}, pages 19045--19053, 2026.

\bibitem[Reichardt et~al.(2026)Reichardt, Ebert, and Wasenm{\"u}ller]{ptx}
Laurenz Reichardt, Nikolas Ebert, and Oliver Wasenm{\"u}ller.
\newblock {PointTransformerX}: Portable and efficient {3D} point cloud processing without sparse algorithms.
\newblock In \emph{CVPRW}, pages 3610--3619, 2026.

\bibitem[Ren et~al.(2026)Ren, Tyszkiewicz, Huang, and Gojcic]{tokengs}
Jiawei Ren, Michal~Jan Tyszkiewicz, Jiahui Huang, and Zan Gojcic.
\newblock {TokenGS}: Decoupling {3D} {G}aussian prediction from pixels with learnable tokens.
\newblock In \emph{CVPR}, pages 15365--15375, 2026.

\bibitem[Rezende and Mohamed(2015)]{normalizingflows}
Danilo~Jimenez Rezende and Shakir Mohamed.
\newblock Variational inference with normalizing flows.
\newblock In \emph{ICML}, pages 1530--1538, 2015.

\bibitem[Rota~Bul{\`o} et~al.(2024)Rota~Bul{\`o}, Porzi, and Kontschieder]{revisingadc}
Samuel Rota~Bul{\`o}, Lorenzo Porzi, and Peter Kontschieder.
\newblock Revising densification in {G}aussian splatting.
\newblock In \emph{ECCV}, pages 347--362, 2024.

\bibitem[Shabanov et~al.(2026)Shabanov, Hedman, Weber, Li, Rozumny, Le~Lan, Dhingra, Luo, Vedaldi, Richardt, Tagliasacchi, Zhu, and Khan]{freerange}
Akhmedkhan Shabanov, Peter Hedman, Ethan Weber, Zhengqin Li, Denis Rozumny, Gael Le~Lan, Naina Dhingra, Lei Luo, Andrea Vedaldi, Christian Richardt, Andrea Tagliasacchi, Bo Zhu, and Numair Khan.
\newblock Free-range {G}aussians: Non-grid-aligned generative {3D} {G}aussian reconstruction.
\newblock In \emph{ECCV}, pages 450--468, 2026.

\bibitem[Sobol'(1967)]{sobol}
Ilya~M. Sobol'.
\newblock On the distribution of points in a cube and the approximate evaluation of integrals.
\newblock \emph{USSR Comput. Math. Math. Phys.}, 7\penalty0 (4):\penalty0 86--112, 1967.

\bibitem[Su et~al.(2024)Su, Ahmed, Lu, Pan, Bo, and Liu]{rope}
Jianlin Su, Murtadha Ahmed, Yu Lu, Shengfeng Pan, Wen Bo, and Yunfeng Liu.
\newblock {RoFormer}: Enhanced transformer with rotary position embedding.
\newblock \emph{Neurocomputing}, 568:\penalty0 127063, 2024.

\bibitem[Tancik et~al.(2020)Tancik, Srinivasan, Mildenhall, Fridovich-Keil, Raghavan, Singhal, Ramamoorthi, Barron, and Ng]{fourier}
Matthew Tancik, Pratul~P. Srinivasan, Ben Mildenhall, Sara Fridovich-Keil, Nithin Raghavan, Utkarsh Singhal, Ravi Ramamoorthi, Jonathan~T. Barron, and Ren Ng.
\newblock {F}ourier features let networks learn high frequency functions in low dimensional domains.
\newblock In \emph{NeurIPS}, pages 7537--7547, 2020.

\bibitem[Veicht et~al.(2026)Veicht, Hong, Bar{\'a}th, and Pollefeys]{zipsplat}
Alexander Veicht, Sunghwan Hong, D{\'a}niel Bar{\'a}th, and Marc Pollefeys.
\newblock {ZipSplat}: Fewer {G}aussians, better splats.
\newblock \emph{arXiv preprint arXiv:2606.05102}, 2026.

\bibitem[Verbin et~al.(2024)Verbin, Srinivasan, Hedman, Mildenhall, Attal, Szeliski, and Barron]{nerfcasting}
Dor Verbin, Pratul~P. Srinivasan, Peter Hedman, Ben Mildenhall, Benjamin Attal, Richard Szeliski, and Jonathan~T. Barron.
\newblock {NeRF-Casting}: Improved view-dependent appearance with consistent reflections.
\newblock In \emph{SIGGRAPH Asia Conference Papers}, pages 1--10, 2024.

\bibitem[Wan et~al.(2026)Wan, Li, Shao, and Zuo]{splatweaver}
Yecong Wan, Fan Li, Mingwen Shao, and Wangmeng Zuo.
\newblock {SplatWeaver}: Learning to allocate {G}aussian primitives for generalizable novel view synthesis.
\newblock \emph{arXiv preprint arXiv:2605.07287}, 2026.

\bibitem[Wang et~al.(2025{\natexlab{a}})Wang, Chen, Karaev, Vedaldi, Rupprecht, and Novotny]{vggt}
Jianyuan Wang, Minghao Chen, Nikita Karaev, Andrea Vedaldi, Christian Rupprecht, and David Novotny.
\newblock {VGGT}: Visual geometry grounded transformer.
\newblock In \emph{CVPR}, pages 5294--5306, 2025{\natexlab{a}}.

\bibitem[Wang et~al.(2025{\natexlab{b}})Wang, Chen, Zhang, Shi, Liu, and Zhuang]{zpressor}
Weijie Wang, Donny~Y. Chen, Zeyu Zhang, Duochao Shi, Akide Liu, and Bohan Zhuang.
\newblock {ZPressor}: Bottleneck-aware compression for scalable feed-forward {3DGS}.
\newblock In \emph{NeurIPS}, pages 125793--125822, 2025{\natexlab{b}}.

\bibitem[Wang et~al.(2024)Wang, Huang, Chen, and Lee]{freesplat}
Yunsong Wang, Tianxin Huang, Hanlin Chen, and Gim~Hee Lee.
\newblock {FreeSplat}: Generalizable {3D} {G}aussian splatting towards free view synthesis of indoor scenes.
\newblock In \emph{NeurIPS}, pages 107326--107349, 2024.

\bibitem[Wu et~al.(2024)Wu, Jiang, Wang, Liu, Liu, Qiao, Ouyang, He, and Zhao]{ptv3}
Xiaoyang Wu, Li Jiang, Peng-Shuai Wang, Zhijian Liu, Xihui Liu, Yu Qiao, Wanli Ouyang, Tong He, and Hengshuang Zhao.
\newblock {P}oint {T}ransformer {V3}: Simpler faster stronger.
\newblock In \emph{CVPR}, pages 4840--4851, 2024.

\bibitem[Xu et~al.(2025)Xu, Peng, Wang, Blum, Barath, Geiger, and Pollefeys]{depthsplat}
Haofei Xu, Songyou Peng, Fangjinhua Wang, Hermann Blum, Daniel Barath, Andreas Geiger, and Marc Pollefeys.
\newblock {DepthSplat}: Connecting {G}aussian splatting and depth.
\newblock In \emph{CVPR}, pages 16453--16463, 2025.

\bibitem[Xu et~al.(2026)Xu, Barath, Geiger, and Pollefeys]{resplat}
Haofei Xu, Daniel Barath, Andreas Geiger, and Marc Pollefeys.
\newblock {ReSplat}: Learning recurrent {G}aussian splatting.
\newblock In \emph{ECCV}, pages 364--382, 2026.

\bibitem[Yan et~al.(2026)Yan, Cao, Wang, Liang, and Guo]{deg}
Runjie Yan, Yan-Pei Cao, Peng Wang, Ding Liang, and Yuan-Chen Guo.
\newblock Generative {3D} {G}aussians with learned density control.
\newblock In \emph{SIGGRAPH Conference Papers}, pages 1--12, 2026.

\bibitem[Yang et~al.(2019)Yang, Huang, Hao, Liu, Belongie, and Hariharan]{pointflow}
Guandao Yang, Xun Huang, Zekun Hao, Ming-Yu Liu, Serge Belongie, and Bharath Hariharan.
\newblock {PointFlow}: {3D} point cloud generation with continuous normalizing flows.
\newblock In \emph{ICCV}, pages 4540--4549, 2019.

\bibitem[Yue et~al.(2026)Yue, Robert, Wang, Hong, Wegner, Rupprecht, and Schindler]{litept}
Yuanwen Yue, Damien Robert, Jianyuan Wang, Sunghwan Hong, Jan~Dirk Wegner, Christian Rupprecht, and Konrad Schindler.
\newblock {LitePT}: Lighter yet stronger point transformer.
\newblock In \emph{CVPR}, pages 24173--24184, 2026.

\bibitem[Zhai et~al.(2025)Zhai, Zhang, Nakkiran, Berthelot, Gu, Zheng, Chen, Bautista, Jaitly, and Susskind]{tarflow}
Shuangfei Zhai, Ruixiang Zhang, Preetum Nakkiran, David Berthelot, Jiatao Gu, Huangjie Zheng, Tianrong Chen, Miguel~{\'A}ngel Bautista, Navdeep Jaitly, and Joshua~M. Susskind.
\newblock Normalizing flows are capable generative models.
\newblock In \emph{ICML}, pages 74348--74369, 2025.

\bibitem[Zhang et~al.(2024{\natexlab{a}})Zhang, Bi, Tan, Xiangli, Zhao, Sunkavalli, and Xu]{gslrm}
Kai Zhang, Sai Bi, Hao Tan, Yuanbo Xiangli, Nanxuan Zhao, Kalyan Sunkavalli, and Zexiang Xu.
\newblock {GS-LRM}: Large reconstruction model for {3D} {G}aussian splatting.
\newblock In \emph{ECCV}, pages 1--19, 2024{\natexlab{a}}.

\bibitem[Zhang et~al.(2024{\natexlab{b}})Zhang, Fei, Liu, Song, and Duan]{ggn}
Shengjun Zhang, Xin Fei, Fangfu Liu, Haixu Song, and Yueqi Duan.
\newblock {G}aussian graph network: Learning efficient and generalizable {G}aussian representations from multi-view images.
\newblock In \emph{NeurIPS}, pages 50361--50380, 2024{\natexlab{b}}.

\bibitem[Zhang et~al.(2026)Zhang, Meng, Wu, and Ding]{sparsesplat}
Zicheng Zhang, Xiangting Meng, Ke Wu, and Wenchao Ding.
\newblock {SparseSplat}: Towards applicable feed-forward {3D} {G}aussian splatting with pixel-unaligned prediction.
\newblock In \emph{CVPR}, pages 5049--5058, 2026.

\bibitem[Zhao et~al.(2021)Zhao, Jiang, Jia, Torr, and Koltun]{pointtransformer}
Hengshuang Zhao, Li Jiang, Jiaya Jia, Philip H.~S. Torr, and Vladlen Koltun.
\newblock {P}oint {T}ransformer.
\newblock In \emph{ICCV}, pages 16259--16268, 2021.

\bibitem[Zhou et~al.(2024)Zhou, Zhang, and Liu]{diffgs}
Junsheng Zhou, Weiqi Zhang, and Yu-Shen Liu.
\newblock {DiffGS}: Functional {G}aussian splatting diffusion.
\newblock In \emph{NeurIPS}, pages 37535--37560, 2024.

\bibitem[Zhou et~al.(2018)Zhou, Tucker, Flynn, Fyffe, and Snavely]{re10k}
Tinghui Zhou, Richard Tucker, John Flynn, Graham Fyffe, and Noah Snavely.
\newblock Stereo magnification: Learning view synthesis using multiplane images.
\newblock \emph{ACM TOG}, 37\penalty0 (4):\penalty0 1--12, 2018.

\bibitem[Ziwen et~al.(2025)Ziwen, Tan, Zhang, Bi, Luan, Hong, Fuxin, and Xu]{longlrm}
Chen Ziwen, Hao Tan, Kai Zhang, Sai Bi, Fujun Luan, Yicong Hong, Li Fuxin, and Zexiang Xu.
\newblock {Long-LRM}: Long-sequence large reconstruction model for wide-coverage {G}aussian splats.
\newblock In \emph{ICCV}, pages 4349--4359, 2025.

\end{thebibliography}
}

\clearpage
\appendix

\onecolumn 
\begin{center}
    {\Large\bfseries \thetitle\par}
    \vspace{0.5em}
    {\large Supplementary Material\par}
\end{center}
\vspace{1em}

\startcontents[supp]
\setcounter{tocdepth}{2}

\noindent{\large\bfseries Contents}\par
\printcontents[supp]{}{1}{}

\section{Method details}

\subsection{Spatial contraction to a bounded domain}
To represent unbounded scenes within the unit-cube domain of the flow,
we follow EGS~\cite{egs} and adopt an $L_\infty$ variant of the
spatial contraction of Mip-NeRF 360~\cite{mipnerf360}.
For a scene position $\mathbf{x}\in\mathbb{R}^3$ after the normalization described in the main paper,
we define $\operatorname{contract}:\mathbb{R}^3\to(-1,1)^3$ as
\begin{equation}
\operatorname{contract}(\mathbf{x}) =
\begin{cases}
a\mathbf{x}, & \|\mathbf{x}\|_\infty \leq 1,\\[6pt]
\left(1-\dfrac{1-a}{\|\mathbf{x}\|_\infty}\right)
\dfrac{\mathbf{x}}{\|\mathbf{x}\|_\infty},
& \text{otherwise},
\end{cases}
\label{eq:contraction}
\end{equation}
where $a=3/8$.
Since the normalization sets the median point distance to one,
at least half of the points observed in the context views lie
in the region where the contraction is a uniform scaling.
To keep every primitive at a finite distance,
we restrict the contracted domain to $[-\kappa,\kappa]^3$ with
$\kappa=0.95$, and map it affinely to $[0,1]^3$.
A position $\boldsymbol{\mu}$ sampled from the flow is
first mapped to contracted coordinates
$\bar{\boldsymbol{\mu}}=\kappa(2\boldsymbol{\mu}-\mathbf{1})$,
and then to normalized scene coordinates through
$\mathbf{x}=\operatorname{uncontract}(\bar{\boldsymbol{\mu}})$, where
\begin{equation}
\operatorname{uncontract}(\bar{\boldsymbol{\mu}}) =
\begin{cases}
\dfrac{\bar{\boldsymbol{\mu}}}{a},
& \|\bar{\boldsymbol{\mu}}\|_\infty \leq a,\\[6pt]
\dfrac{(1-a)\bar{\boldsymbol{\mu}}}
{\|\bar{\boldsymbol{\mu}}\|_\infty
(1-\|\bar{\boldsymbol{\mu}}\|_\infty)},
& \text{otherwise}.
\end{cases}
\end{equation}

Following EGS~\cite{egs}, we apply the distance-dependent correction of
NeRF-Casting~\cite{nerfcasting} to the predicted Gaussian scales, multiplying them by
$\rho^2/(2\rho-1)$, where $\rho=\max(\|\mathbf{x}\|_2,1)$.
This allows a given predicted scale to result in a similar screen-space
footprint at any distance from the scene center.

\subsection{Rational-quadratic spline transformations}
Each flow block transforms one coordinate at a time using a
monotonic rational-quadratic spline~\cite{splines} defined over $[0,1]$.
The spline consists of $M=16$ segments connecting knots
$\{(x^{(m)},y^{(m)})\}_{m=0}^{M}$, from $(0,0)$ to $(1,1)$,
with positive derivatives $\delta^{(m)}$ at the knots.
For an input $z$ in segment $m$, we define the segment width
$w_m=x^{(m+1)}-x^{(m)}$, height $h_m=y^{(m+1)}-y^{(m)}$,
slope $\sigma_m=h_m/w_m$, and the relative position
$\xi=(z-x^{(m)})/w_m$.
The transformation is given by
\begin{equation}
f(z)=y^{(m)}+
\frac{h_m\left[
\sigma_m\xi^2+\delta^{(m)}\xi(1-\xi)
\right]}
{\sigma_m+
\left[\delta^{(m+1)}+\delta^{(m)}-2\sigma_m\right]
\xi(1-\xi)}.
\label{eq:rqs}
\end{equation}
The inverse is obtained in closed form by solving a quadratic equation,
while the analytical derivative is used for log-density evaluation.

For each coordinate, the network predicts $3M+1$
unconstrained parameters corresponding to the segment widths,
heights, and knot derivatives.
Unlike the original formulation~\cite{splines}, our bounded
domain requires no linear tails, allowing us to learn
the derivatives at both endpoints.
We parameterize these values as
\begin{equation}
\begin{aligned}
w_m &= \epsilon+(1-M\epsilon)
    \operatorname{softmax}(\boldsymbol{\eta}^{w})_m,\\
h_m &= \epsilon+(1-M\epsilon)
    \operatorname{softmax}(\boldsymbol{\eta}^{h})_m,\\
\delta^{(m)} &= \epsilon+
    \frac{1}{\beta}\log(1+e^{\beta\eta_m^\delta}),
\end{aligned}
\end{equation}
where $\epsilon=10^{-3}$ and
$\beta=\log 2/(1-\epsilon)$.
The minimum widths, heights, and derivatives prevent
degenerate segments and keep the log-derivatives finite.
We initialize the output layer of each spline head to zero,
yielding uniform segments and unit derivatives, so that
each flow block starts as the identity transformation and the flow initially 
samples uniformly from the domain.

\subsection{Score-function gradient implementation}
We implement the score-function estimator through an opacity surrogate,
replacing the opacity of each primitive with
\begin{equation}
\tilde{o}_i =
o_i \exp\left(
\log p_\theta(\boldsymbol{\mu}_i\mid\mathcal{C})
-
\operatorname{sg}\left(
\log p_\theta(\boldsymbol{\mu}_i\mid\mathcal{C})
\right)
\right),
\end{equation}
where $\operatorname{sg}$ denotes the stop-gradient operator.
During the forward pass, $\tilde{o}_i=o_i$, leaving the rendered image unchanged.
We detach the sampled positions from the flow to prevent pathwise gradients,
so that gradients with respect to the flow parameters propagate only
through the surrogate opacities, yielding
\begin{equation}
\begin{aligned}
\nabla_\theta \mathcal{L}
&= \sum_{i=1}^{N}
\frac{\partial\mathcal{L}}{\partial\tilde{o}_i}
\nabla_\theta\tilde{o}_i\\
&= \sum_{i=1}^{N}
o_i\frac{\partial\mathcal{L}}{\partial o_i}
\nabla_\theta\log p_\theta(
\boldsymbol{\mu}_i\mid\mathcal{C}).
\end{aligned}
\end{equation}
This recovers the score-function gradient using standard automatic differentiation,
requiring no modifications to the rasterizer.

\subsection{Inference-time sampling}
During training, we draw the base samples of the flow independently
and uniformly.
At inference, we instead use a scrambled Sobol sequence~\cite{sobol,scrambling}.
This provides more uniform coverage of the base domain and slightly
improves reconstruction quality, particularly at low primitive counts,
as shown in \cref{tab:qmc_sampling}.

We also use this sampling procedure for progressive local refinement.
To refine a specific region of an existing reconstruction,
we identify the primitives in the selected region and recover their
corresponding base samples by inverting the flow.
We draw additional base samples using the same Sobol sampling procedure,
retaining those whose nearest original base sample belongs to the
selection, until the desired number of new samples is reached.
The retained samples are transformed through the flow to obtain new
primitive positions.
We then apply the attribute network to the combined set of existing
and new primitives, but retain its predictions only for the primitives in the selected region,
leaving the other primitives unchanged.
\begin{table}[h]
\centering
\footnotesize
\caption{\textbf{Quasi-random sampling at inference.} Difference between scrambled Sobol and independent uniform base samples (Sobol minus uniform) 
for our method on the 4-view DL3DV-Evaluation interpolation cases of QuerySplat~\cite{querysplat}, averaged over five sampling seeds for each variant. 
Positive $\Delta$PSNR and $\Delta$SSIM and negative $\Delta$LPIPS favor Sobol sampling. \#G denotes the number of Gaussian primitives per scene.}
\label{tab:qmc_sampling}
\setlength{\tabcolsep}{4pt}
\begin{tabular}{l ccc}
\toprule
\#G & $\Delta$PSNR$\uparrow$ & $\Delta$SSIM$\uparrow$ & $\Delta$LPIPS$\downarrow$ \\
\midrule
8K & +0.34 & +0.009 & $-$0.008 \\
32K & +0.14 & +0.003 & $-$0.004 \\
128K & +0.06 & +0.001 & $-$0.001 \\
512K & +0.03 & +0.001 & $-$0.001 \\
1M & +0.02 & +0.001 & $-$0.001 \\
\bottomrule
\end{tabular}
\end{table}

\section{Additional comparisons}

\subsection{Cross-dataset generalization}
We evaluate cross-dataset generalization on RealEstate10K~\cite{re10k}
and ACID~\cite{acid}, using the 8-view evaluation indices of
ZPressor~\cite{zpressor}~(\cref{tab:generalization_mv}).
All methods use their DL3DV-trained checkpoints without further
fine-tuning, and are evaluated at $256\times448$ resolution.
Across both datasets, \paper generalizes well and continues to
benefit from larger inference-time sampling budgets.
On RealEstate10K, our method already surpasses all baselines in PSNR at 57K primitives,
and achieves the best performance across all three metrics
at 262K primitives, with further gains at 524K.
This includes baselines that use significantly more
primitives and are initialized from RealEstate10K-trained models.
On ACID, our model achieves higher PSNR than the baselines at the same primitive budget.
\begin{table}[t]
\centering
\footnotesize
\caption{
    \textbf{Cross-dataset generalization}
    on RealEstate10K~\cite{re10k} and ACID~\cite{acid},
    using the 8-view evaluation indices of ZPressor~\cite{zpressor} at $256\times448$.
    $^\ddagger$Initialized from a model trained on RealEstate10K.
    {\setlength{\fboxsep}{1pt}
    \colorbox{tableone}{\textbf{Best}},
    \colorbox{tabletwo}{\underline{second-best}}, and
    \colorbox{tablethree}{third-best}}
    results are highlighted accordingly.
}
\label{tab:generalization_mv}
\setlength{\tabcolsep}{3pt}
\begin{tabular}{l cccc cccc}
\toprule
& \multicolumn{4}{c}{RealEstate10K}
& \multicolumn{4}{c}{ACID} \\
\cmidrule(lr){2-5}
\cmidrule(lr){6-9}
Method
& \#G & PSNR$\uparrow$ & SSIM$\uparrow$ & LPIPS$\downarrow$
& \#G & PSNR$\uparrow$ & SSIM$\uparrow$ & LPIPS$\downarrow$ \\
\midrule
\multicolumn{9}{l}{\textit{Pixel-aligned}} \\
DepthSplat~\cite{depthsplat}$^\ddagger$
& 918K & 25.14 & 0.866 & 0.141
& 918K & 26.01 & 0.810 & \sbest\underline{0.184} \\

IDESplat~\cite{idesplat}$^\ddagger$
& 918K & 25.60 & 0.872 & \tbest 0.139
& 918K & 26.67 & \tbest 0.822 & \best\textbf{0.172} \\

ReSplat (init)~\cite{resplat}
& 57K & 27.34 & 0.888 & 0.141
& 57K & 26.52 & 0.816 & 0.205 \\

\midrule
\multicolumn{9}{l}{\textit{Query-based}} \\
TokenGS~\cite{tokengs}
& 262K & 26.65 & 0.865 & 0.203
& 262K & 27.30 & 0.814 & 0.281 \\

LocusGS~\cite{locusgs}
& 262K & 27.48 & \tbest 0.891 & 0.165
& 262K & \tbest 27.93 & \best\textbf{0.837} & 0.245 \\

\midrule
\multirow{3}{*}{Ours}
& 57K & \tbest 27.62 & 0.872 & 0.157
& 57K & 26.77 & 0.788 & 0.239 \\

& 262K & \sbest\underline{28.94} & \sbest\underline{0.897} & \sbest\underline{0.129}
& 262K & \sbest\underline{28.05} & \sbest\underline{0.825} & 0.198 \\

& 524K & \best\textbf{29.34} & \best\textbf{0.904} & \best\textbf{0.121}
& 524K & \best\textbf{28.48} & \best\textbf{0.837} & \tbest 0.185 \\
\bottomrule
\end{tabular}
\end{table}

\subsection{Extrapolation on QuerySplat cases}
In \cref{tab:qs_extrap}, we report novel-view extrapolation results on the DL3DV-Evaluation
cases of QuerySplat~\cite{querysplat} across 2, 4, and 12 input views.
Consistent with the interpolation results in the main paper,
with 66K primitives, \paper outperforms all baselines in PSNR and SSIM
across the three input-view settings.
Increasing the budget to 262K further improves reconstruction quality,
giving the best results across all three metrics and view counts.
\begin{table*}[htbp]
\centering
\footnotesize
\setlength{\tabcolsep}{3.5pt}
\caption{\textbf{Extrapolation results} of novel-view synthesis on the DL3DV-Evaluation cases of QuerySplat~\cite{querysplat}, averaged over small, medium, and large splits at $256\times448$ resolution.
{\setlength{\fboxsep}{1pt}
\colorbox{tableone}{\textbf{Best}},
\colorbox{tabletwo}{\underline{second-best}}, and
\colorbox{tablethree}{third-best}}
results are highlighted accordingly.}
\label{tab:qs_extrap}
\begin{tabular}{l cccc cccc cccc}
\toprule
& \multicolumn{4}{c}{2 views}
& \multicolumn{4}{c}{4 views}
& \multicolumn{4}{c}{12 views} \\
\cmidrule(lr){2-5}
\cmidrule(lr){6-9}
\cmidrule(lr){10-13}
Method
& \#G & PSNR$\uparrow$ & SSIM$\uparrow$ & LPIPS$\downarrow$
& \#G & PSNR$\uparrow$ & SSIM$\uparrow$ & LPIPS$\downarrow$
& \#G & PSNR$\uparrow$ & SSIM$\uparrow$ & LPIPS$\downarrow$ \\
\midrule

\multicolumn{13}{l}{\textit{Pixel-aligned}} \\

DepthSplat~\cite{depthsplat}
& 229K & 19.77 & 0.690 & \tbest 0.257
& 459K & 18.98 & 0.675 & \sbest\underline{0.276}
& 1.38M & 16.16 & 0.558 & \sbest\underline{0.384} \\

IDESplat~\cite{idesplat}
& 229K & 20.07 & \tbest 0.691 & 0.263
& 459K & 19.23 & 0.685 & \tbest 0.278
& 1.38M & 16.31 & 0.552 & 0.403 \\

ReSplat (init)~\cite{resplat}
& 14K & 19.65 & 0.665 & 0.324
& 29K & 19.03 & 0.683 & 0.301
& 86K & 16.57 & \tbest 0.586 & \tbest 0.385 \\

\midrule
\multicolumn{13}{l}{\textit{Query-based}} \\

TokenGS~\cite{tokengs}
& 262K & 20.10 & 0.658 & 0.390
& 262K & 19.89 & 0.651 & 0.396
& 262K & 17.10 & 0.530 & 0.507 \\

LocusGS~\cite{locusgs}
& 262K & 20.71 & 0.685 & 0.357
& 262K & \tbest 20.78 & \tbest 0.694 & 0.352
& 262K & \tbest 18.00 & 0.582 & 0.460 \\

\midrule

\multirow{3}{*}{Ours}
& 14K & \tbest 20.78 & 0.677 & 0.333
& 14K & 19.82 & 0.644 & 0.363
& 14K & 17.22 & 0.535 & 0.464 \\

& 66K & \sbest\underline{22.13} & \sbest\underline{0.737} & \sbest\underline{0.253}
& 66K & \sbest\underline{21.28} & \sbest\underline{0.708} & 0.282
& 66K & \sbest\underline{18.32} & \sbest\underline{0.592} & 0.390 \\

& 262K & \best\textbf{22.69} & \best\textbf{0.760} & \best\textbf{0.224}
& 262K & \best\textbf{22.05} & \best\textbf{0.742} & \best\textbf{0.245}
& 262K & \best\textbf{19.03} & \best\textbf{0.631} & \best\textbf{0.347} \\

\bottomrule
\end{tabular}
\end{table*}

\subsection{Square-frame comparisons}
QuerySplat~\cite{querysplat} reconstructs only the central square of each image,
and it estimates its own cameras from the input views.
To compare against it, we evaluate all methods under its evaluation protocol.
Every method receives only the central square of each input image,
resized to $256\times256$ ($512\times512$ for QuerySplat),
and every method other than QuerySplat receives the dataset cameras.
The central square of each target view is rendered and scored at $256\times256$
following the released evaluation code of QuerySplat.
We report interpolation and extrapolation results
in~\cref{tab:qs_square_interp,tab:qs_square_extrap}.
With 66K primitives, \paper outperforms QuerySplat across all metrics and view counts,
using eight times fewer primitives.
\begin{table}[h]
\centering
\footnotesize
\caption{\textbf{Square-frame interpolation results} on DL3DV-Evaluation, under the evaluation protocol of QuerySplat~\cite{querysplat}.
$^\ast$QuerySplat estimates its own cameras from the images, while every other method receives the dataset cameras.
{\setlength{\fboxsep}{1pt}
\colorbox{tableone}{\textbf{Best}},
\colorbox{tabletwo}{\underline{second-best}}, and
\colorbox{tablethree}{third-best}}
results are highlighted accordingly.}
\label{tab:qs_square_interp}
\setlength{\tabcolsep}{3pt}
\begin{tabular}{l cccc cccc cccc}
\toprule
& \multicolumn{4}{c}{2 views}
& \multicolumn{4}{c}{4 views}
& \multicolumn{4}{c}{12 views} \\
\cmidrule(lr){2-5}
\cmidrule(lr){6-9}
\cmidrule(lr){10-13}
Method
& \#G & PSNR$\uparrow$ & SSIM$\uparrow$ & LPIPS$\downarrow$
& \#G & PSNR$\uparrow$ & SSIM$\uparrow$ & LPIPS$\downarrow$
& \#G & PSNR$\uparrow$ & SSIM$\uparrow$ & LPIPS$\downarrow$ \\
\midrule
\multicolumn{13}{l}{\textit{Pixel-aligned}} \\
DepthSplat~\cite{depthsplat}
& 131K & 19.76 & 0.659 & 0.294
& 262K & 22.52 & 0.768 & 0.207
& 786K & 21.97 & 0.773 & 0.211 \\

IDESplat~\cite{idesplat}
& 131K & 20.25 & 0.674 & 0.289
& 262K & 23.05 & 0.783 & 0.204
& 786K & 22.08 & 0.774 & 0.226 \\

ReSplat (init)~\cite{resplat}
& 8K & 18.56 & 0.620 & 0.362
& 16K & 21.35 & 0.737 & 0.265
& 49K & 23.61 & \tbest 0.806 & \tbest 0.203 \\

\midrule
\multicolumn{13}{l}{\textit{Query-based}} \\
TokenGS~\cite{tokengs}
& 262K & 20.37 & 0.671 & 0.362
& 262K & 23.65 & 0.770 & 0.269
& 262K & 22.42 & 0.721 & 0.327 \\

LocusGS~\cite{locusgs}
& 262K & 20.69 & 0.687 & 0.336
& 262K & 24.82 & \tbest 0.812 & 0.224
& 262K & 23.90 & 0.781 & 0.267 \\

QuerySplat~\cite{querysplat}$^\ast$
& 524K & 21.39 & 0.699 & \tbest 0.259
& 524K & 24.57 & 0.800 & \tbest 0.184
& 524K & 23.70 & 0.768 & 0.230 \\

\midrule
\multirow{3}{*}{Ours}
& 14K & \tbest 21.99 & \tbest 0.724 & 0.285
& 14K & \tbest 24.83 & 0.804 & 0.218
& 14K & \tbest 24.25 & 0.779 & 0.250 \\

& 66K & \sbest\underline{22.93} & \sbest\underline{0.757} & \sbest\underline{0.240}
& 66K & \sbest\underline{26.77} & \sbest\underline{0.849} & \sbest\underline{0.161}
& 66K & \sbest\underline{26.73} & \sbest\underline{0.849} & \sbest\underline{0.170} \\

& 262K & \best\textbf{23.15} & \best\textbf{0.765} & \best\textbf{0.230}
& 262K & \best\textbf{27.40} & \best\textbf{0.865} & \best\textbf{0.144}
& 262K & \best\textbf{27.89} & \best\textbf{0.880} & \best\textbf{0.135} \\

\bottomrule
\end{tabular}
\end{table}

\begin{table}[h]
\centering
\footnotesize
\caption{\textbf{Square-frame extrapolation results} on DL3DV-Evaluation, under the evaluation protocol of QuerySplat~\cite{querysplat}.
$^\ast$QuerySplat estimates its own cameras from the images, while every other method receives the dataset cameras.
{\setlength{\fboxsep}{1pt}
\colorbox{tableone}{\textbf{Best}},
\colorbox{tabletwo}{\underline{second-best}}, and
\colorbox{tablethree}{third-best}}
results are highlighted accordingly.}
\label{tab:qs_square_extrap}
\setlength{\tabcolsep}{3pt}
\begin{tabular}{l cccc cccc cccc}
\toprule
& \multicolumn{4}{c}{2 views}
& \multicolumn{4}{c}{4 views}
& \multicolumn{4}{c}{12 views} \\
\cmidrule(lr){2-5}
\cmidrule(lr){6-9}
\cmidrule(lr){10-13}
Method
& \#G & PSNR$\uparrow$ & SSIM$\uparrow$ & LPIPS$\downarrow$
& \#G & PSNR$\uparrow$ & SSIM$\uparrow$ & LPIPS$\downarrow$
& \#G & PSNR$\uparrow$ & SSIM$\uparrow$ & LPIPS$\downarrow$ \\
\midrule
\multicolumn{13}{l}{\textit{Pixel-aligned}} \\
DepthSplat~\cite{depthsplat}
& 131K & 19.78 & 0.689 & 0.273
& 262K & 19.22 & 0.686 & 0.285
& 786K & 16.58 & 0.576 & 0.391 \\

IDESplat~\cite{idesplat}
& 131K & 20.25 & 0.699 & 0.271
& 262K & 19.22 & 0.696 & 0.286
& 786K & 16.68 & 0.575 & 0.404 \\

ReSplat (init)~\cite{resplat}
& 8K & 18.89 & 0.649 & 0.339
& 16K & 18.26 & 0.662 & 0.330
& 49K & 16.36 & 0.588 & 0.391 \\

\midrule
\multicolumn{13}{l}{\textit{Query-based}} \\
TokenGS~\cite{tokengs}
& 262K & 20.83 & 0.691 & 0.338
& 262K & 20.77 & 0.695 & 0.339
& 262K & 17.70 & 0.572 & 0.458 \\

LocusGS~\cite{locusgs}
& 262K & 21.16 & 0.705 & 0.315
& 262K & \tbest 21.56 & \tbest 0.733 & 0.304
& 262K & 18.58 & \tbest 0.622 & 0.416 \\

QuerySplat~\cite{querysplat}$^\ast$
& 524K & 21.38 & 0.703 & \tbest 0.252
& 524K & 21.34 & 0.715 & \tbest 0.263
& 524K & \tbest 18.80 & 0.618 & \tbest 0.370 \\

\midrule
\multirow{3}{*}{Ours}
& 14K & \tbest 21.87 & \tbest 0.729 & 0.285
& 14K & 20.90 & 0.699 & 0.314
& 14K & 17.97 & 0.583 & 0.421 \\

& 66K & \sbest\underline{23.13} & \sbest\underline{0.771} & \sbest\underline{0.230}
& 66K & \sbest\underline{22.37} & \sbest\underline{0.749} & \sbest\underline{0.254}
& 66K & \sbest\underline{19.16} & \sbest\underline{0.635} & \sbest\underline{0.359} \\

& 262K & \best\textbf{23.56} & \best\textbf{0.786} & \best\textbf{0.213}
& 262K & \best\textbf{23.07} & \best\textbf{0.775} & \best\textbf{0.228}
& 262K & \best\textbf{19.87} & \best\textbf{0.668} & \best\textbf{0.326} \\

\bottomrule
\end{tabular}
\end{table}

\subsection{Comparisons under the DepthSplat protocol}
To broaden the comparison, we also report results under the DepthSplat protocol.
We include author-reported results for methods that have not released
checkpoints for this setting.
Baselines with released checkpoints are evaluated by us.
As shown in~\cref{tab:depthsplat_bench}, our model achieves the best
results across all metrics and view counts with 262K primitives.
With smaller budgets of 33K, 66K, and 98K for the respective view counts,
it also outperforms all baselines while using fewer primitives
than ATSplat, the strongest baseline in the four- and six-view settings.
\begin{table*}[htbp]
\centering
\footnotesize
\caption{
    \textbf{Comparison under the DepthSplat protocol}~\cite{depthsplat} on DL3DV at $256\times448$ resolution.
    $^\dagger$Author-reported results. The remaining baselines are evaluated by us.
    {\setlength{\fboxsep}{1pt}
    \colorbox{tableone}{\textbf{Best}},
    \colorbox{tabletwo}{\underline{second-best}}, and
    \colorbox{tablethree}{third-best}}
    results are highlighted accordingly.
}
\label{tab:depthsplat_bench}
\setlength{\tabcolsep}{3pt}
\begin{tabular}{l cccc cccc cccc}
\toprule
& \multicolumn{4}{c}{2 views}
& \multicolumn{4}{c}{4 views}
& \multicolumn{4}{c}{6 views} \\
\cmidrule(lr){2-5}
\cmidrule(lr){6-9}
\cmidrule(lr){10-13}
Method
& \#G & PSNR$\uparrow$ & SSIM$\uparrow$ & LPIPS$\downarrow$
& \#G & PSNR$\uparrow$ & SSIM$\uparrow$ & LPIPS$\downarrow$
& \#G & PSNR$\uparrow$ & SSIM$\uparrow$ & LPIPS$\downarrow$ \\
\midrule

\multicolumn{13}{l}{\textit{Pixel-aligned}} \\

DepthSplat~\cite{depthsplat}
& 229K & 19.61 & 0.626 & 0.299
& 459K & 23.15 & 0.779 & 0.175
& 688K & 24.15 & 0.818 & 0.146 \\

IDESplat~\cite{idesplat}
& 229K & 19.95 & 0.637 & \cellcolor{tablethree}0.298
& 459K & 23.57 & 0.790 & 0.174
& 688K & 24.66 & 0.830 & 0.145 \\

ReSplat (init)~\cite{resplat}
& 14K & 18.75 & 0.579 & 0.400
& 29K & 23.93 & 0.782 & 0.213
& 43K & 25.71 & 0.835 & 0.167 \\

iLRM~\cite{ilrm}$^\dagger$
& -- & -- & -- & --
& -- & -- & -- & --
& 172K & 25.60 & 0.830 & 0.168 \\

\midrule
\multicolumn{13}{l}{\textit{Query-based}} \\

TokenGS~\cite{tokengs}
& 262K & 19.90 & 0.616 & 0.434
& 262K & 22.71 & 0.723 & 0.338
& 262K & 23.56 & 0.750 & 0.318 \\

LocusGS~\cite{locusgs}
& 262K & 20.35 & 0.644 & 0.400
& 262K & 24.02 & 0.781 & 0.273
& 262K & 24.91 & 0.805 & 0.251 \\

\midrule
\multicolumn{13}{l}{\textit{Adaptive (pixel-aligned)}} \\

ATSplat~\cite{atsplat}$^\dagger$
& 40K & 20.07 & 0.630 & 0.332
& 81K
& \cellcolor{tablethree}25.67
& \cellcolor{tablethree}0.827
& \cellcolor{tablethree}0.172
& 121K
& \cellcolor{tablethree}27.28
& \cellcolor{tablethree}0.868
& \cellcolor{tablethree}0.138 \\

SparseSplat~\cite{sparsesplat}$^\dagger$
& -- & -- & -- & --
& -- & -- & -- & --
& 150K & 24.20 & 0.817 & 0.168 \\

\midrule
\multirow{3}{*}{Ours}
& 14K
& \cellcolor{tablethree}21.44
& \cellcolor{tablethree}0.671
& 0.327
& 29K & 25.00 & 0.798 & 0.211
& 43K & 26.55 & 0.842 & 0.170 \\

& 33K
& \cellcolor{tabletwo}\underline{22.09}
& \cellcolor{tabletwo}\underline{0.703}
& \cellcolor{tabletwo}\underline{0.282}
& 66K
& \cellcolor{tabletwo}\underline{25.98}
& \cellcolor{tabletwo}\underline{0.832}
& \cellcolor{tabletwo}\underline{0.170}
& 98K
& \cellcolor{tabletwo}\underline{27.59}
& \cellcolor{tabletwo}\underline{0.871}
& \cellcolor{tabletwo}\underline{0.135} \\

& 262K
& \cellcolor{tableone}\textbf{22.87}
& \cellcolor{tableone}\textbf{0.737}
& \cellcolor{tableone}\textbf{0.235}
& 262K
& \cellcolor{tableone}\textbf{26.97}
& \cellcolor{tableone}\textbf{0.861}
& \cellcolor{tableone}\textbf{0.134}
& 262K
& \cellcolor{tableone}\textbf{28.44}
& \cellcolor{tableone}\textbf{0.893}
& \cellcolor{tableone}\textbf{0.111} \\

\bottomrule
\end{tabular}
\end{table*}

\section{Additional ablations}
In~\cref{tab:ablations_supp}, we provide additional ablations of the flow depth,
attribute network depth, attention window size, and reward clamping using the same training setup
as the ablations in the main paper.
Reducing the attribute network from twelve to six layers lowers
PSNR by 0.40\,dB, while reducing the flow from five to three blocks
has a smaller effect.
Halving the attention window from 256 to 128 primitives also causes a small reduction in quality.
The largest drop comes from removing reward clamping, which we examine below.
\begin{table}[h]
\centering
\footnotesize
\setlength{\tabcolsep}{4pt}
\caption{\textbf{Additional ablations} on DL3DV-Evaluation using the 4-view interpolation cases of QuerySplat~\cite{querysplat}, with 110K Gaussian primitives per scene. The full model uses an attribute network with 12 layers and attention windows of 256 primitives, a normalizing flow with 5 blocks, and the clamped estimator. The best result is shown in \textbf{bold}.}
\label{tab:ablations_supp}
\begin{tabular}{lccc}
\toprule
Variant & PSNR$\uparrow$ & SSIM$\uparrow$ & LPIPS$\downarrow$ \\
\midrule
Ours (full model) & \textbf{24.87} & \textbf{0.805} & \textbf{0.195} \\
\midrule
AttrNet 9 layers & 24.74 & 0.801 & 0.198 \\
AttrNet 6 layers & 24.47 & 0.794 & 0.206 \\
\midrule
AttrNet window 128 & 24.83 & 0.803 & 0.196 \\
\midrule
Flow 4 blocks & 24.84 & 0.804 & 0.197 \\
Flow 3 blocks & 24.79 & 0.802 & 0.200 \\
\midrule
Ours w/ full estimator & 21.11 & 0.667 & 0.325 \\
\bottomrule
\end{tabular}
\end{table}

\subsection{Reward clamping analysis}
As shown in~\cref{tab:ablations_supp}, removing reward
clamping and training with the full score-function estimator (retaining
both positive and negative opacity gradients) causes a severe drop in reconstruction quality.
To understand this behavior, we measure the gradient signal-to-noise ratio
(SNR) of the density estimator with respect to the flow parameters at 10K, 20K,
and 30K training iterations (\cref{tab:estimator_snr}).
We define the SNR as the inverse of the simplified gradient noise scale~\cite{gradientsnr},
$\operatorname{SNR} = \|\mathbb{E}[\mathbf{g}]\|_2^2 / \operatorname{tr}(\operatorname{Cov}(\mathbf{g}))$,
where $\mathbf{g}$ is the flow-gradient estimate.
We use 100 independent estimates, each from 110K primitives,
and average over eight DL3DV validation scenes.
Indeed, across all three stages of training, the SNR of the full estimator
sits at least two orders of magnitude lower than that of the clamped estimator.
While introducing bias to the estimator, clamping the reward significantly reduces the
noise of the gradient and results in better reconstruction quality.
\begin{table}[h]
\centering
\footnotesize
\caption{\textbf{Gradient signal-to-noise ratio of the score-function estimator.} SNR is reported for the model trained with the clamped estimator, and for a model trained with the full estimator. Each estimator is evaluated and averaged over 8 held-out scenes with 100 estimates per scene. Ratio: clamped over full.}
\label{tab:estimator_snr}
\setlength{\tabcolsep}{4pt}
\begin{tabular}{l ccc}
\toprule
Iteration & Clamped & Full & Ratio \\
\midrule
10K & 35.1 & $<$\,0.01 & $>$\,1000$\times$ \\
20K & 23.5 & 0.11     & $\sim$\,210$\times$ \\
30K & 11.0 & $<$\,0.01 & $>$\,1000$\times$ \\
\bottomrule
\end{tabular}
\end{table}

\section{Further analysis}

\subsection{Geometry evaluation}
To evaluate geometric accuracy and completeness, we follow
Surflo~\cite{surflo} and reconstruct reference meshes from all
scene frames using Gaussian Wrapping~\cite{gaussianwrapping}.
We sample 10M surface points per mesh, with density inversely
proportional to the squared distance to the nearest capture camera.
For each case, we retain points visible from any frame in the
interpolation window.
Visibility is determined by comparing projected point depths
with rendered mesh depths, using a 1\% relative tolerance.

All methods use a budget of 28,672 primitives, matching ReSplat~\cite{resplat}.
Our model samples this count directly, while larger baseline outputs are randomly subsampled.
Both point sets are then voxelized at $0.1\%$ of $d$, where $d$
is the diagonal of the bounding box of the visible reference points.

We use Surflo's evaluation code to compute accuracy, completeness,
Chamfer distance, precision, recall, and F1.
Distances are reported in units of $10^{-2}d$,
and precision and recall use a threshold of $0.01d$.
All metrics are computed per case and averaged.

\subsection{Computational efficiency}
We compare inference runtime and peak GPU memory with
TokenGS~\cite{tokengs}, LocusGS~\cite{locusgs}, and
QuerySplat~\cite{querysplat} using four input views on a single B200 GPU (\cref{tab:compute}).
We separate image encoding, which runs once per scene,
from primitive generation, and further break down our generation stage into
the normalizing flow and attribute network.
The variable budgets for the query-based baselines are obtained by resizing their learned queries, and runtimes are the median of ten runs.
At the same primitive budget, \paper is slower than TokenGS and LocusGS, but faster than QuerySplat.

Our coordinate quantization substantially reduces the cost
of the normalizing flow.
The Transformer in each flow block is evaluated once per scene
on its discretized grid, leaving only feature interpolation
and spline transformations to be computed per primitive.
At 1M primitives, this reduces the flow runtime from
4.3 seconds to 78\,ms, approximately a 55-fold speedup.
Because it operates per primitive, the attribute network accounts for most
of the generation time and peak memory use at large budgets.
Its attention operates within fixed-size windows, so its
computational cost scales linearly with the primitive count.
A lighter image encoder and greater sharing of computation
across primitives during attribute prediction could
further reduce the runtime and memory costs.
\begin{table}[h]
\centering
\footnotesize
\caption{\textbf{Inference runtime and peak GPU memory} on a single NVIDIA B200 GPU with four input views, excluding rasterization.}
\label{tab:compute}
\setlength{\tabcolsep}{3.5pt}
\begin{tabular}{l c cc cc cc}
\toprule
& & \multicolumn{2}{c}{66K}
& \multicolumn{2}{c}{262K}
& \multicolumn{2}{c}{1M} \\
\cmidrule(lr){3-4}
\cmidrule(lr){5-6}
\cmidrule(lr){7-8}
Method & Encoder (ms)
& Generation (ms) & Mem. (GB)
& Generation (ms) & Mem. (GB)
& Generation (ms) & Mem. (GB) \\
\midrule
TokenGS~\cite{tokengs}
& 4
& 7 & 1.3
& 12 & 1.4
& 49 & 1.8 \\

LocusGS~\cite{locusgs}
& 4
& 18 & 1.6
& 35 & 2.2
& 130 & 4.6 \\

QuerySplat~\cite{querysplat}
& 227
& 64 & 8.8
& 275 & 9.1
& 1522 & 10.4 \\

\midrule
Ours
& 79
& 52 & 7.3
& 105 & 11.0
& 355 & 22.3 \\

\quad \textit{Normalizing flow}
& --
& 27 & 6.2
& 34 & 7.8
& 78 & 10.0 \\

\quad \textit{Attribute network}
& --
& 24 & 7.3
& 72 & 11.0
& 277 & 22.3 \\

\midrule
\quad \textit{Normalizing flow w/o quantization}
& --
& 268 & --
& 1068 & --
& 4274 & -- \\

\bottomrule
\end{tabular}
\end{table}

\clearpage
\section{Additional qualitative results}
\begin{figure*}[h]
    \centering
    \includegraphics[width=\textwidth]{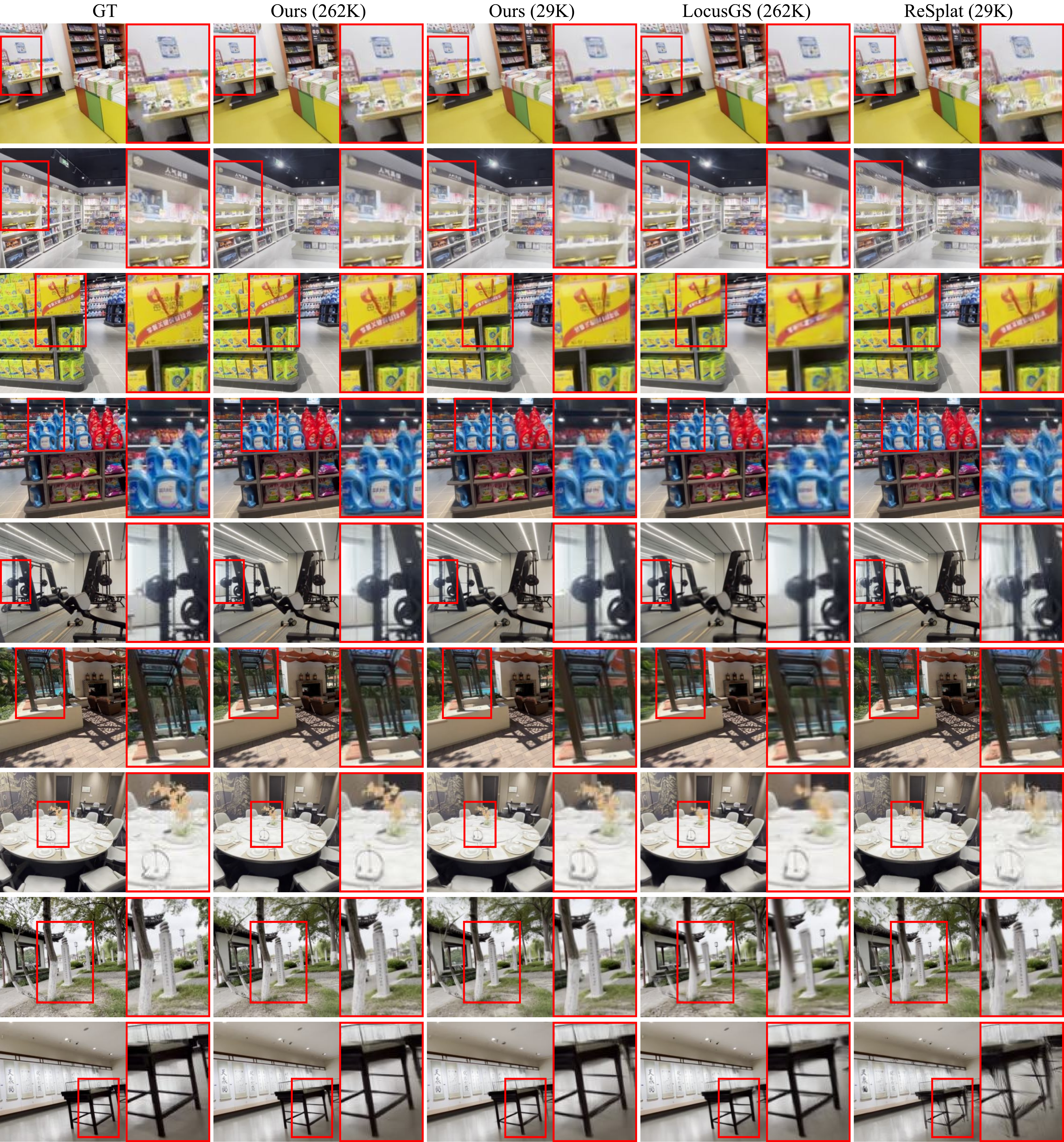} 
    \caption{
        \textbf{Qualitative comparison} on DL3DV-Evaluation using 4 views. The number of Gaussians is indicated next to each method.
    }
    \label{fig:quali_supp}
\end{figure*}

\end{document}